\documentclass[10pt,twocolumn,letterpaper]{article}

\usepackage{cvpr}              
\definecolor{cvprblue}{rgb}{0.21,0.49,0.74}
\usepackage[pagebackref,breaklinks,colorlinks,allcolors=cvprblue]{hyperref}
\usepackage{booktabs}
\usepackage{multirow}
\usepackage{graphicx}
\usepackage{adjustbox}

\def\paperID{***} 
\def\confName{CVPR}
\def\confYear{2026}

\title{Thinking Diffusion: Penalize and Guide Visual-Grounded Reasoning in Diffusion Multimodal Language Models}

\author{
Keuntae Kim$^{1}$\thanks{Equal contribution.} \quad
Mingyu Kang$^{2}$\footnotemark[1] \quad
Yong Suk Choi$^{1,2}$\thanks{Corresponding author.} \\
$^{1}$Department of Computer Science, Hanyang University \\
$^{2}$Department of Artificial Intelligence, Hanyang University \\
{\tt\small ktkpv94@hanyang.ac.kr, alsrb15788@hanyang.ac.kr, cys@hanyang.ac.kr}
}

\begin{document}
\maketitle
\begin{abstract}
Diffusion large language models (dLLMs) are emerging as promising alternatives to autoregressive (AR) LLMs. Recently, this paradigm has been extended to multimodal tasks, leading to the development of diffusion multimodal large language models (dMLLMs). These models are expected to retain the reasoning capabilities of LLMs while enabling faster inference through parallel generation. However, when combined with Chain-of-Thought (CoT) reasoning, dMLLMs exhibit two critical issues.
First, we observe that dMLLMs often generate the final answer token at a very early timestep. This trend indicates that the model determines the answer before sufficient reasoning, leading to degraded reasoning performance. Second, during the initial timesteps, dMLLMs show minimal dependency on visual prompts, exhibiting a fundamentally different pattern of visual information utilization compared to AR vision-language models. In summary, these findings indicate that dMLLMs tend to generate premature final answers without sufficiently grounding on visual inputs.
To address these limitations, we propose Position \& Step Penalty (PSP) and Visual Reasoning Guidance (VRG). PSP penalizes tokens in later positions during early timesteps, delaying premature answer generation and encouraging progressive reasoning across timesteps. VRG, inspired by the classifier-free guidance, amplifies visual grounding signals to enhance the model’s alignment with visual evidence.
Extensive experiments across various dMLLMs demonstrate that our method achieves up to 7.5\% higher accuracy while delivering more than 3× speedup compared to reasoning with four times more diffusion steps.
\end{abstract}    
\section{Introduction}
\label{sec:intro}

Vision-language models (VLMs) have demonstrated remarkable capabilities in understanding visual inputs and generating contextually relevant textual descriptions \cite{vlmsurvey, vlm1_falmingo, vlm2_blip2, vlm3_llava, vlm4_intervl, vlm5_qwen2vl}. These models have achieved strong performance across a wide range of vision-language tasks, including visual question answering and visual reasoning. Furthermore, with the incorporation of Chain-of-Thought (CoT) prompting, VLM can perform step-by-step reasoning before producing the final answer \cite{vlm_cot1, vlm_cot2, ccot}.

Recently, diffusion large language models (dLLMs) \cite{llada, dream} have emerged as a promising alternative to autoregressive (AR) LLMs. In contrast to AR models that generate a single token at a time in a left-to-right manner, dLLMs restore multiple tokens in parallel, providing substantially faster inference. Extending this advantage to the multimodal model, diffusion multimodal large language models (dMLLMs) \cite{mllm2, mllm3, mmada, lavida} have been introduced, presenting a new paradigm that enables joint reasoning over textual and visual information.

\begin{figure*}[t]
  \centering
  \includegraphics[width=1.0\linewidth]{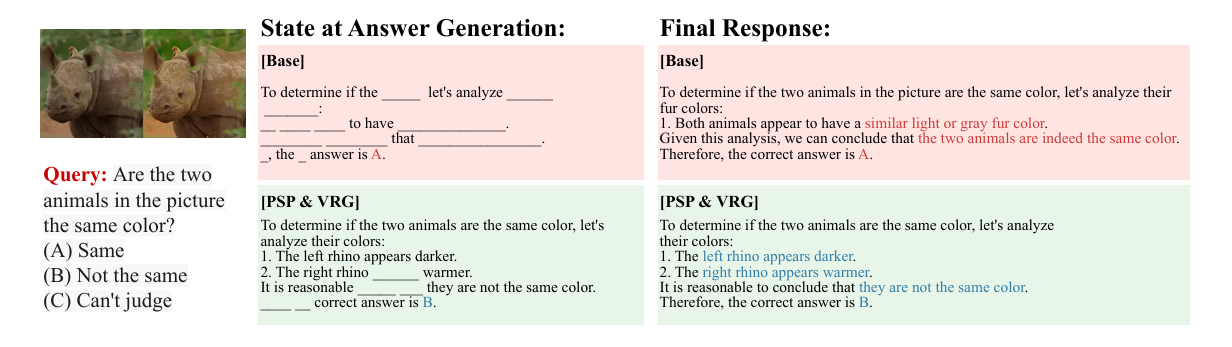}
  \caption{Example responses of LaViDa (generation length = 64, diffusion steps = 32). \textbf{State at answer generation} refers to the output state at the moment when the final answer (A, B, C) is generated, while \textbf{Final Response} denotes the model’s final response generated by each method.}
  \label{fig:casestudy}
\end{figure*}

However, the reasoning process of dMLLMs remains insufficiently understood. Since dMLLMs rely on a diffusion-based generation mechanism that reconstructs tokens in parallel, reasoning enhancement methods designed for AR VLMs cannot be directly applied. Moreover, while prior research has provided quantitative analyses on how AR VLMs utilize visual evidence \cite{favero2024multi, jiang2025devils, jung2025visual}, such analyses are largely absent for dMLLMs, leaving it unclear whether they effectively leverage visual information. This gap highlights the need to closely investigate how dMLLMs perform reasoning and to design effective methods to strengthen their reasoning ability.

In this work, we present the first quantitative analysis of the reasoning process of dMLLMs with CoT prompting. Our analysis presents two critical issues. First, dMLLMs exhibit Early Answer Generation, where the model generates final answer tokens prematurely at very early timesteps, particularly when the number of diffusion steps is small. For example, as shown in Figure \ref{fig:casestudy}, the model appears to determine the final answer before completing sufficient reasoning steps and then generates intermediate reasoning to justify the final answer. Second, we observe low dependency on visual prompts during the initial timesteps. Unlike AR VLMs, dMLLMs exhibit a fundamentally different pattern, relying little on visual information early on and incorporating visual cues only in later timesteps. Overall, these findings suggest that dMLLMs tend to generate answers too early without effectively leveraging visual evidence.

To address these issues, we propose two novel training-free methods that can be applied directly during inference: Position \& Step Penalty (PSP) and Visual Reasoning Guidance (VRG).
First, PSP discourages the premature generation of answer-position tokens at early timesteps by applying penalties, thereby encouraging a more progressive reasoning process. Second, VRG, inspired by classifier-free guidance, amplifies the conditional logits associated with visual prompts to strengthen the model’s use of visual evidence.

We validate the effectiveness of our method through extensive experiments. Experimental results across various dMLLMs demonstrate that our method achieves up to 7.5\% higher accuracy and over 3× faster inference compared to reasoning using four times more diffusion steps. Our contributions can be summarized as follows:

\begin{itemize}\itemsep 4pt
\item We conduct the first quantitative analysis of the reasoning process in diffusion-based multimodal language models with CoT prompting, identifying Early Answer Generation and insufficient early visual grounding.

\item We propose Position \& Step Penalty (PSP) to promote timestep-wise reasoning progression and Visual Reasoning Guidance (VRG) to enhance visual grounding during inference.

\item We introduce a training-free method that generalizes across different dMLLMs and achieves strong performance on multimodal benchmarks, outperforming models that use four times more diffusion steps with up to 7.5\% higher accuracy and more than 3× faster inference.
\end{itemize}

\begin{figure*}
  \centering
  \begin{subfigure}{0.48\linewidth}
    \includegraphics[width=\textwidth]{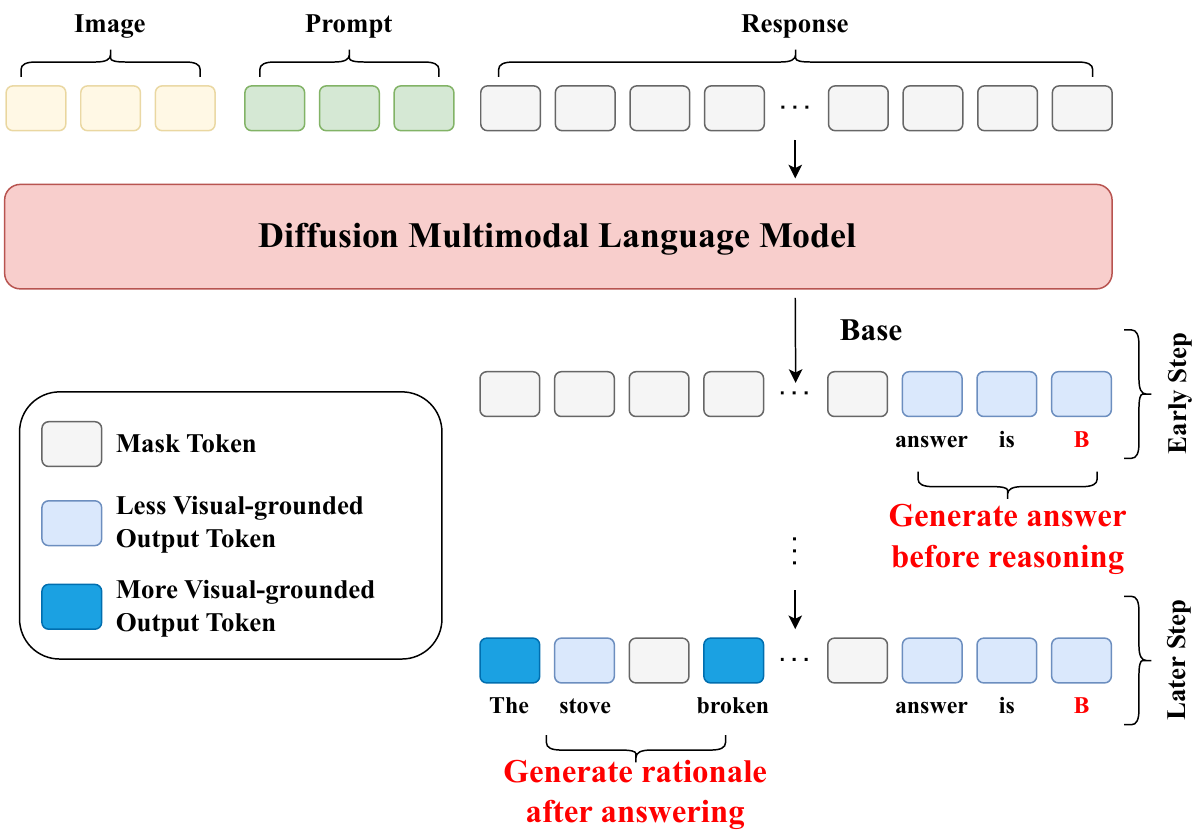}
    \caption{Baseline dMLLM Inference.}
    \label{fig:overview_a}
  \end{subfigure}
  \hfill
  \begin{subfigure}{0.48\linewidth}
    \includegraphics[width=\textwidth]{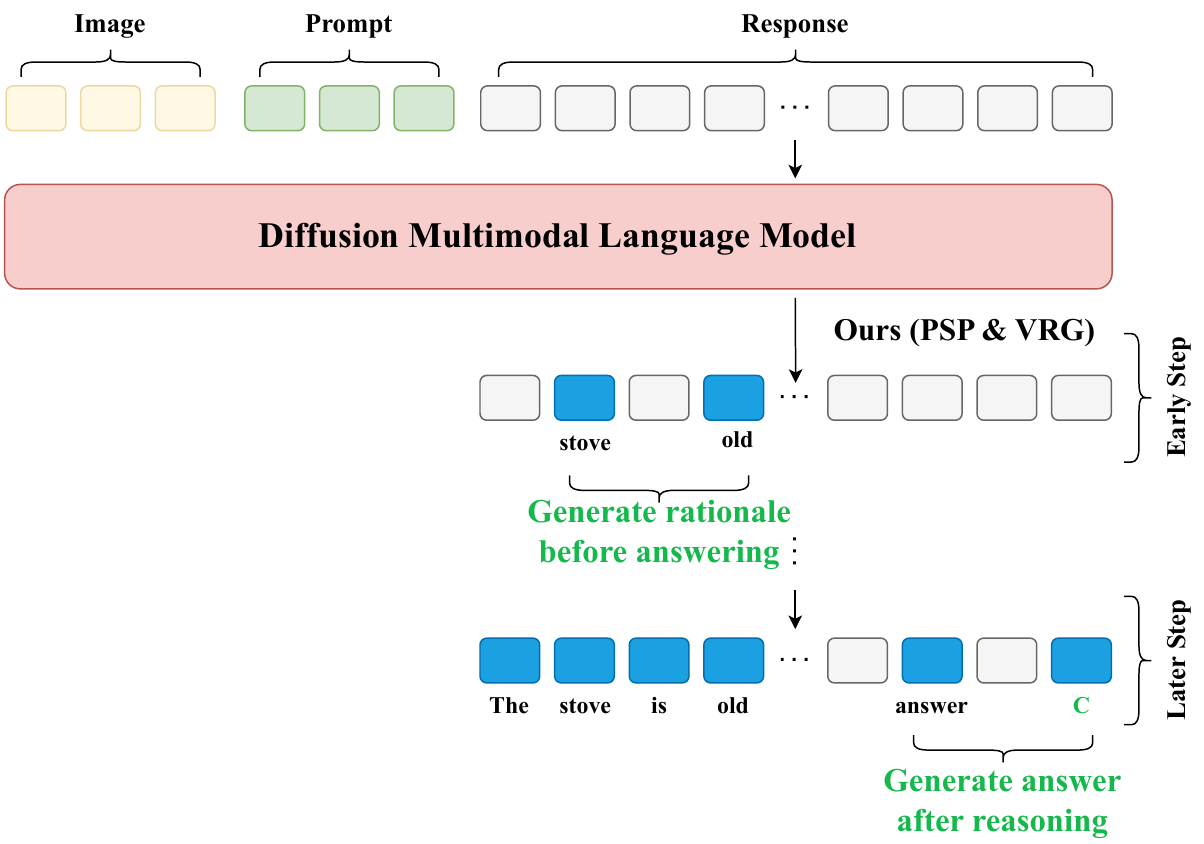}
    \caption{Our proposed dMLLM inference.}
    \label{fig:overview_b}
  \end{subfigure}
  \caption{Overview of our method. (a) shows the base inference process, which exhibits early answer generation and less visual grounding. (b) illustrates our proposed inference process that introduces PSP and VRG to mitigate these two issues.}
  \label{fig:overview}
\end{figure*}

\section{Background \& Related Work}
\label{sec:related_work}

\subsection{Diffusion Language Models}
\label{subsec:dlm}

Diffusion models have demonstrated notable success in image generation \cite{rombach2022high, ramesh2022hierarchical}. Motivated by their success, several studies have adapted diffusion methods to text generation by applying them to continuous text embeddings \cite{li2022diffusion, gong2022diffuseq, dieleman2022continuous}. To handle the discrete nature of text, discrete diffusion models were introduced, operating directly on the discrete vocabulary space \cite{sahoo2024simple, lou2310discrete}. Recently, dLLMs such as LLaDA \cite{llada} and Dream \cite{dream} achieved performance comparable to autoregressive LLMs at scale. These models allow flexible control over the speed–quality tradeoff by adjusting the number of diffusion steps.


Formally, given a discrete token sequence of length $L$, $X_0 = [X^1_0, X^2_0, \ldots, X^L_0]$, the forward process $q(X_t | X_s)$ progressively masks tokens over the time interval $[0, 1]$, where $1 \geq t \geq s \geq 0$. When $t = 1$, the sequence $X_1$ consists entirely of mask tokens [$M$]. The model $p_\theta$ parameterizes the reverse process $p(X_s | X_t)$, learning to reconstruct the original sequence from the fully masked state.

During inference, the model begins with a fully masked sequence $X_1 = [M, M, \ldots, M]$ and iteratively applies the learned reverse process $p_\theta (X_0 | X_t)$ to gradually restore the original tokens. Sampling starts by setting the target generation length and initializing the response sequence $X_1$ entirely with [$M$]. The model then progressively transitions from state $X_t$ to $X_s$ (where $s < t$), decreasing the masking level and incrementally recovering the sequence.

Specifically, given the current state $X_t$, the model $p_\theta$ predicts all masked tokens [$M$] based on the provided conditions (e.g., multimodal input $v$, prompt $c$, and the partially masked sequence $X_t$). After prediction, a portion of tokens corresponding to the ratio $s/t$ is remasked, while the remaining $(1 - s/t)$ tokens are retained as is.

\subsection{Multimodal Chain-of-Thought}
\label{subsec:mutimodal_cot}
Chain-of-Thought (CoT) \cite{cot1, cot2, cot3}, which has greatly contributed to improving the reasoning ability of LLMs, has also been actively explored in MLLMs \cite{multimodal_cot2, multimodal_cot4, multimodal_cot3}. However, unlike CoT in LLMs, Multimodal CoT faces the challenge of handling inputs from different modalities, and various approaches have been proposed to address this issue \cite{multimodal_cot}. The most common strategy is to leverage the strong capabilities of LLMs by generating image captions and feeding them, together with textual prompts, as inputs to the LLM. However, this approach heavily depends on the quality of the image captions and suffers from significant visual information loss during the image-to-text conversion process.

To overcome these limitations, there have been attempts to apply multimodal CoT directly to VLMs. Nevertheless, such methods still rely heavily on text-based rationales and struggle to capture fine-grained associations between visual and textual information \cite{vlm_cot1,vlm_cot2, vlm_cot3}. Recent studies have thus focused on enriching visual descriptions or injecting structural information. For instance, CCoT \cite{ccot} generates a scene graph to structurally encode object, relation, and spatial information into the prompt, thereby systematizing linguistic reasoning cues. DDCoT \cite{ddcot} decomposes complex inputs into reasoning and recognition and guides the model through a step-by-step reasoning process. ICoT \cite{icot} attempts to insert fine visual cues, which are difficult to express in text, directly into intermediate reasoning steps by interleaving textual rationales with image patches. Although these methods have proven effective in autoregressive (AR) VLMs, they fundamentally depend on stepwise reasoning through sequential token generation and therefore fail to demonstrate the same effectiveness in dMLLMs.

\subsection{CoT in Diffusion Language Models}
\label{subsec:cot_dlm}

In AR based language models, CoT method has achieved remarkable success by generating intermediate reasoning steps \cite{cot1, cot2, cot3}. AR CoT leverages the left-to-right token generation process, conditioning each new token on previously generated ones. This approach maximizes reasoning capability by allowing the model to generate and utilize its own rationales. However, it also has critical drawbacks. As the length of the intermediate reasoning increases, the inference time and computational cost of AR CoT grow linearly. Moreover, due to the irreversible generation structure, once a token is generated, it cannot be modified. Consequently, any error that occurs during generation continues to propagate through subsequent timesteps. To correct such errors, the framework must force the model to regenerate the sequence from the beginning, which requires additional inference calls and results in several times higher computational cost \cite{llm_frame1, llm_frame2, llm_frame3, llm_frame4}.

In contrast, the reasoning process of dLLMs consists of iterative steps, where tokens are unmasked and the remaining tokens are remasked for progressive refinement at each timestep \cite{diffusion_cot1, dlm_infer1}. Thus, the core of diffusion CoT lies in the remasking strategy \cite{dlm_infer2, dlm_infer3}. While intuitive strategies such as low-confidence, entropy, and margin-based methods have been explored, research on optimal remasking strategies for reasoning remains in its early stages. Although diffusion CoT has not yet reached the reasoning performance level of AR CoT, it offers two key advantages that AR CoT cannot provide: (1) faster reasoning through parallel generation, (2) flexible control over the speed–quality tradeoff. In this work, we extend dMLLMs with CoT reasoning while effectively balancing the speed–quality tradeoff inherent in diffusion-based generation.


\begin{figure}[t]
  \centering
  \begin{subfigure}{\columnwidth}
    \includegraphics[width=\textwidth]{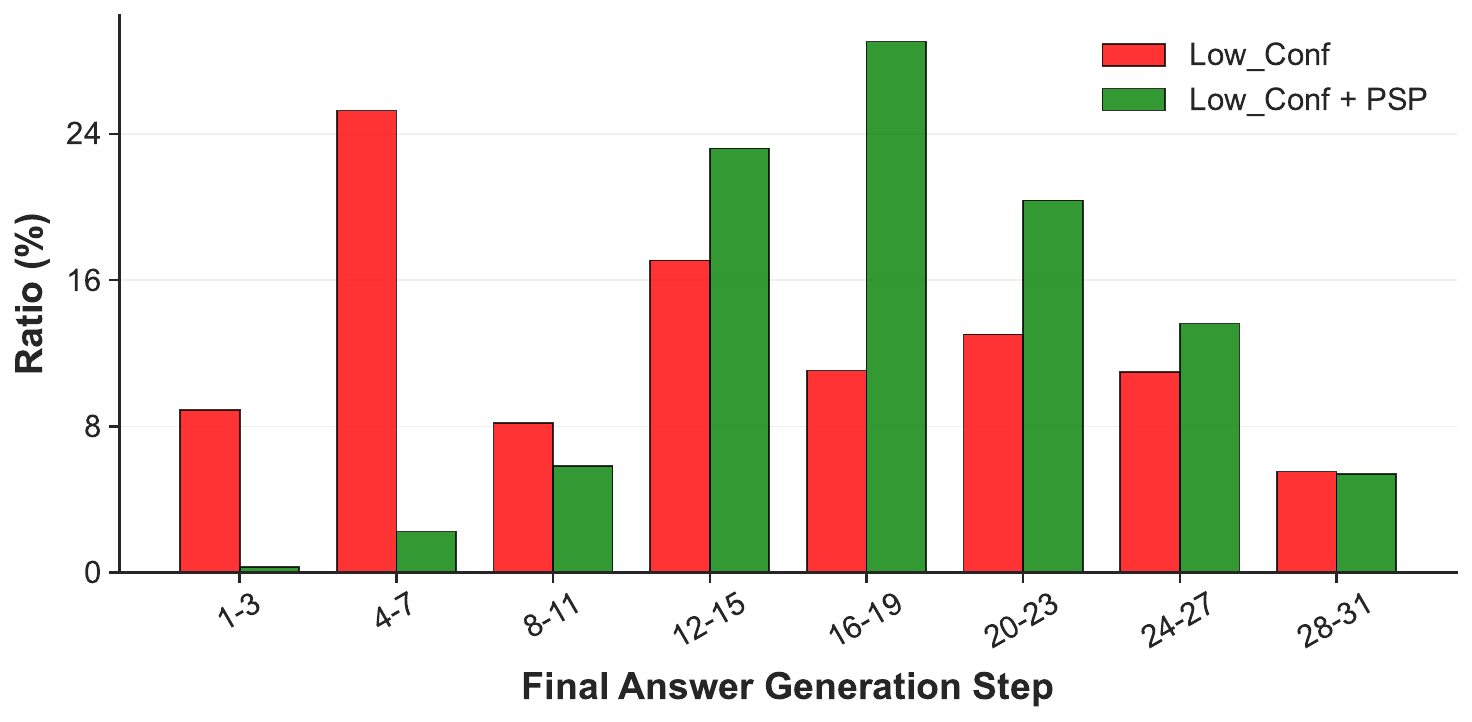}
    \caption{Generation length $L$ = 64 / diffusion step $T$ =32}
    \label{fig:answer_steps_64}
  \end{subfigure}
  \vspace{0mm} 
  \begin{subfigure}{\columnwidth}
    \includegraphics[width=\textwidth]{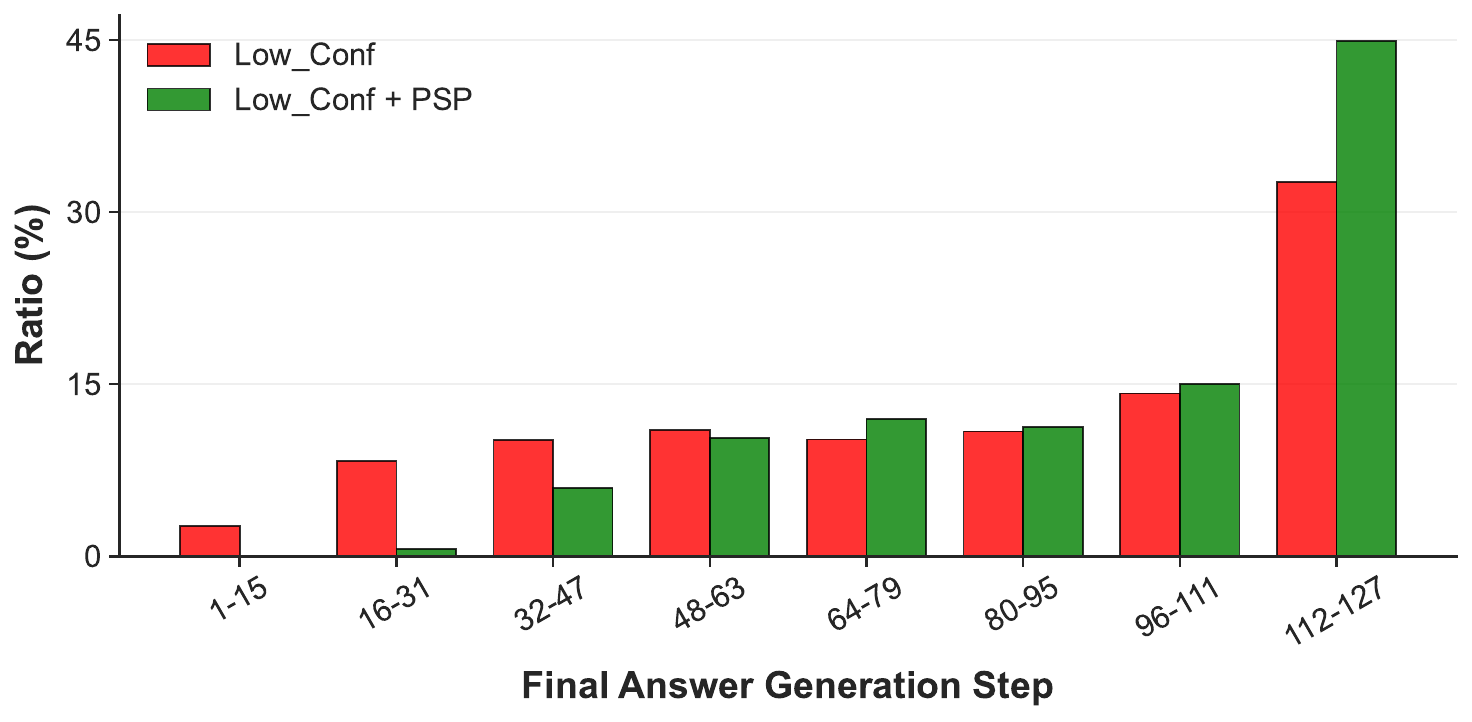}
    \caption{Generation length $L$ = 256 / diffusion step $T$ = 128}
    \label{fig:answer_steps_256}
  \end{subfigure}
  \caption{Results of the final answer generation step on the M3CoT validation set using LaViDa. The default remasking strategy is Low-confidence.}
  \label{fig:answer_step}
\end{figure}

\section{Analysis of Reasoning in dMLLMs}
\label{sec:analysis}

\subsection{Experimental Setup for Case Study}
\label{subsec:casestudy_setup}
To analyze the reasoning of dMLLMs with CoT prompting, we conduct a case study using M3CoT Validation Set \cite{m3cot}. M3CoT covers diverse domains such as science, mathematics, and commonsense, requiring multi-step reasoning based on multimodal inputs. We quantitatively analyze the inference process of dMLLMs under varying diffusion steps and response generation length. Our primary analysis focuses on LaViDa, while additional case studies on other dMLLMs are provided in the supplementary material.

\subsection{Observation 1: Early Answer Generation}
\label{subsec:casestudy1}
We quantitatively analyze when dMLLMs generate an answer under different diffusion steps $T$ and generation length $L$. Specifically, for each sample, we record the timestep at which the final answer token first appears and visualize the overall distribution in Figure \ref{fig:answer_step}.

Figure \ref{fig:answer_steps_64} shows the distribution under $L$ = 64 and $T$ = 32. In this setting, the model tends to generate the final answer at very early steps. Specifically, it exhibits a strong Early Answer Generation tendency, producing the final answer before the 7th step in over 30\% of cases, as shown in Figure \ref{fig:answer_steps_64}.
This suggests that it often generates the answer before sufficient reasoning has been completed, and only afterward generates rationales based on that early answer. In contrast, as shown in the Figure \ref{fig:answer_steps_256}, with $L$ = 256 and $T$ = 128, most samples generate answers at later steps. This indicates that when both the diffusion steps and generation length are sufficiently large, the model tends to complete a more coherent reasoning process before generating the final answer.

Overall, these results suggest that while stable reasoning emerges with sufficiently large $L$ and $T$, smaller configurations lead to a stronger Early Answer Generation, preventing the model from performing sufficient reasoning. Therefore, to ensure reliable reasoning performance even under limited $L$ and $T$, a new remasking strategy is required to guide the model’s reasoning progression more effectively.

\subsection{Observation 2: Week Visual Grounding}
\label{subsec:casestudy2}
We further evaluate how visual information contributes to token generation. Following prior work \cite{favero2024multi}, we quantify the dependency of each unmasked token on visual prompts using the visual prompt dependency measure (PDM):

\begin{equation}
\resizebox{0.90\linewidth}{!}{$
\text{PDM}(X_s)
= \frac{1}{\sqrt{2}}
\sqrt{
  \sum \left(
    \sqrt{p_\theta(X_s \mid X_t, c, v)}
    - \sqrt{p_\theta(X_s \mid X_t, c)}
  \right)^2
}
$}
\end{equation}

\noindent
where $v$, $c$, and $X_t$ denote the visual input, the prompt, and the partially masked sequence, respectively.

\begin{figure}[t]
  \centering
  \begin{subfigure}{\columnwidth}
    \includegraphics[width=\textwidth]{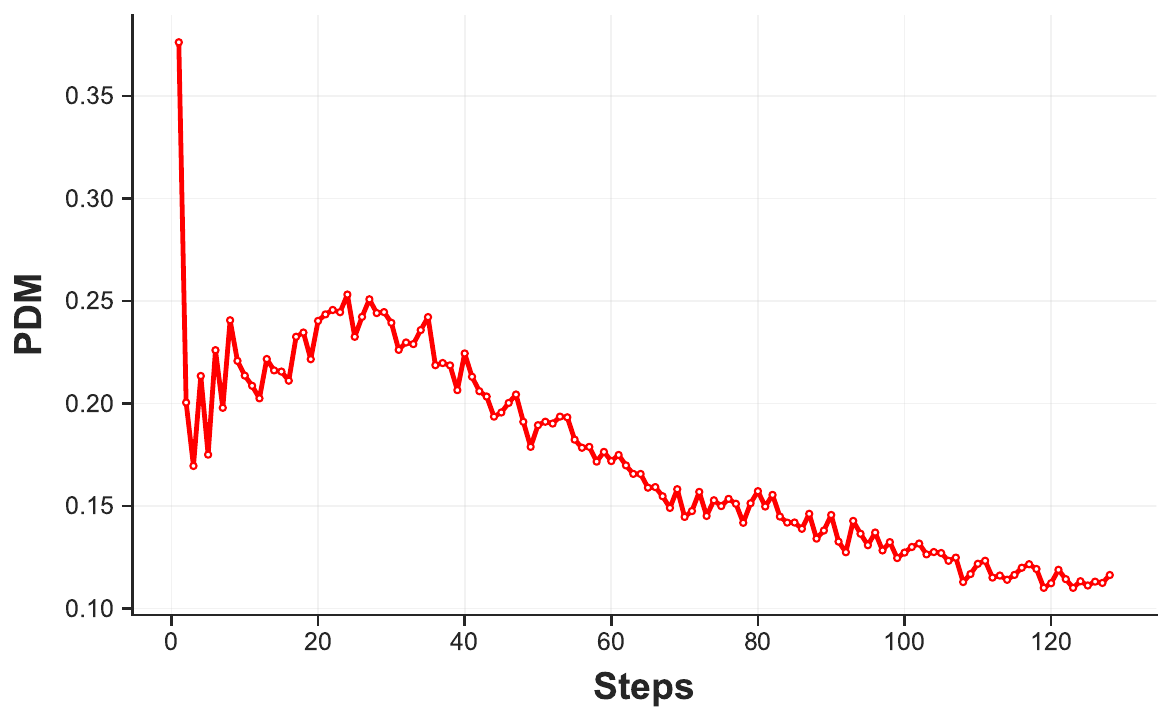}
    \caption{LLaVA-1.5}
    \label{fig:llava_pdm}
  \end{subfigure}
  \vspace{3mm} 
  \begin{subfigure}{\columnwidth}
    \includegraphics[width=\textwidth]{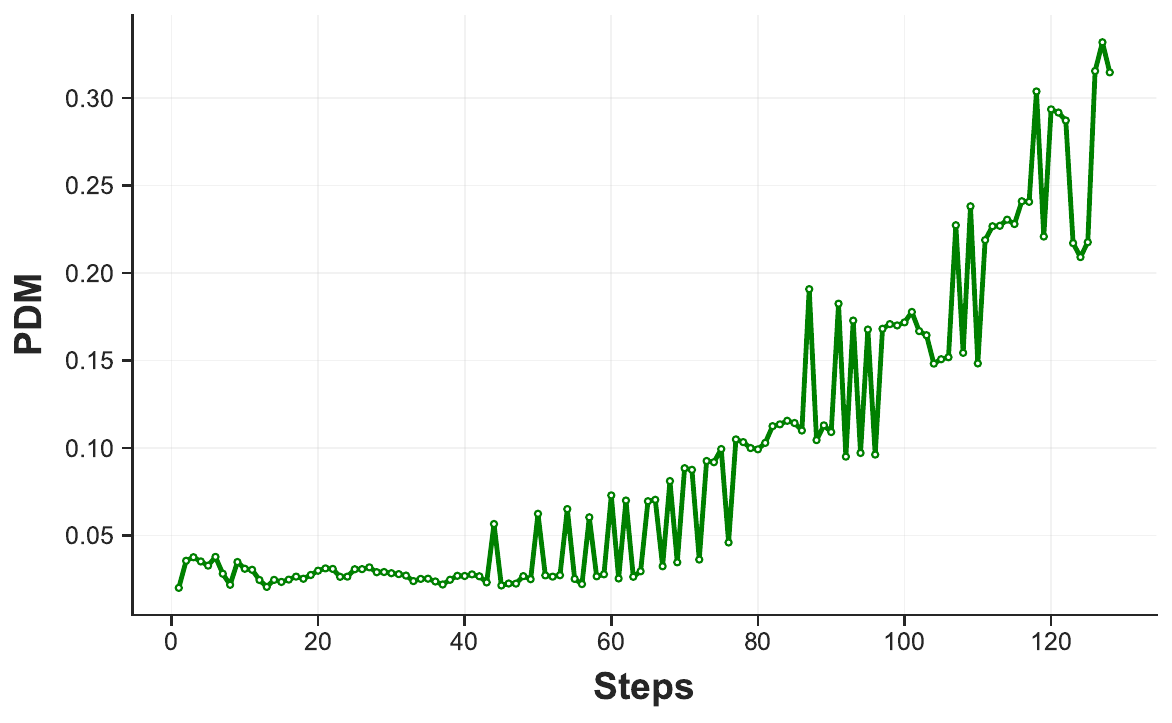}
    \caption{LaVida with generation length $L$ = 256 / step $T$ = 128}
    \label{fig:lavida_pdm}
  \end{subfigure}
  \caption{Comparison of PDM measurements on the M3CoT validation set between the autoregressive-based model LLaVA-1.5 and the diffusion-based model LaViDa.}
  \label{fig:llava_vs_lavida}
\end{figure}

As shown in Figure~\ref{fig:llava_vs_lavida}, LLaVA-1.5 \cite{liu2024improved} (an AR-based VLM) presents high PDM at the early stages of generation, which gradually decline as reasoning progresses. This trend suggests that the model heavily relies on visual features at the beginning but reduces its dependence on them in later reasoning steps \cite{favero2024multi, jung2025visual}. In contrast, dMLLM shows a markedly different pattern. The overall PDM are lower, particularly at the early diffusion steps, where the model shows minimal sensitivity to visual inputs. As the diffusion process progresses, PDM gradually increases, indicating that the model begins to incorporate visual information only at later stages of generation.

This result suggests that, unlike AR VLMs that actively incorporate visual information from the earliest reasoning steps, dMLLMs generate their final answer already at very early timesteps, before visual prompt dependence has been sufficiently strengthened. Combining Observations 1 and 2, we find that dMLLMs often generate an answer without sufficiently referencing the visual input and then construct the following reasoning trajectory around that answer, which is generated before effective visual grounding.

\section{Method}
\label{sec:method}

\subsection{Position \& Step Penalty}
\label{subsec:PSP}
In Section \ref{subsec:casestudy1}, we observe that dMLLMs tend to generate answers prematurely under short diffusion steps and limited response generation length. In other words, the model often generates the final answer tokens before completing sufficient reasoning steps, and subsequently generates rationales conditioned on this early answer.

To mitigate this issue, we introduce Position \& Step Penalty (PSP), designed to delay answer generation to later timesteps and encourage gradual reasoning progression. As shown in Figure \ref{fig:overview_b}, the model starts from an initial sequence $X_1$ consisting of $L$ masked tokens and progressively unmasks them through $K$ discrete timesteps $\{t_1, t_2, \ldots, t_K\}$, where $t_1 = 1$ and $t_K = 0$. At each step, the model samples:

\begin{equation}
    X_0 \sim p_\theta(X_0 \mid X_{t_i})
\end{equation}
and then remasks the top $L \times t_{i+1}$ tokens to obtain $X_{t_{i+1}}$.

For each $j$th token at timestep $t_i$, with its confidence score $C_j^{i}$, we apply PSP as follows:

\begin{equation}
\mathrm{rel}(j)\in[0,1], \qquad 
\tau_i = \frac{i}{K} \in (0,1]
\end{equation}

\begin{equation}
    \tilde{C}_j^{i} = C_j^{i} \cdot \left[ 1 - \gamma (1 - \tau_i) \, \text{rel}(j) \right]
\label{eq:4}
\end{equation}
where $\gamma$ is a penalty strength coefficient, $\tau_i$ represents the relative progress of the diffusion process, and $\text{rel}(j)$ denotes the normalized positional index of token $j$ within the response sequence (ranging from 0 to 1). Consequently, tokens positioned toward the end of the sequence receive a much stronger penalty in the early timesteps, preventing the model from prematurely generating an answer token. This penalty encourages the model to establish intermediate reasoning steps before generating the final answer.

As shown in Figure \ref{fig:answer_step}, applying PSP effectively shifts the answer generation step to later timesteps compared to the baseline. This indicates that the model performs more complete reasoning before generating the final answer.

\begin{table*}[htbp]
    \centering
    \caption{Main result table. X/Y denotes the generation length $L$ and step $T$, respectively. \textit{Low-conf} refers to the Low-confidence remasking strategy, and \textit{Ours} indicates the Low-conf strategy combined with PSP and VRG. \textbf{Bold} represents the best performance, and \underline{underline} indicates the second-best performance.}
    \label{tab:main}
    \begin{adjustbox}{max width=\textwidth}
    \begin{tabular}{l|l|ccc|ccc|ccc|ccc}
        \toprule
        \multirow{2}{*}{\textbf{Model}} & \multirow{2}{*}{\textbf{Method}} &
        \multicolumn{3}{c}{\textbf{M$^3$CoT}} & \multicolumn{3}{c}{\textbf{MMBench}} & \multicolumn{3}{c}{\textbf{SQA-IMG}} & \multicolumn{3}{c}{\textbf{V* Bench}} \\
        \cmidrule(lr){3-5} \cmidrule(lr){6-8} \cmidrule(lr){9-11} \cmidrule(lr){12-14}
        & & 64/32 & 128/64 & 256/128 & 64/32 & 128/64 & 256/128 & 64/32 & 128/64 & 256/128 & 64/32 & 128/64 & 256/128  \\
        \midrule

        \multirow{7}{*}{\textit{LaViDa}} 
        & DDCoT & 45.7 & 46.7 & 46.7 & 72.7 & 72.8 & 73.7 & 71.1 & 71.1 & 71.7 & 41.3 & 42.4 & 43.9 \\
        & CCoT & 45.2 & 46.5 & 47.7 & 72.8 & 73.9 & 73.4 & 71.2 & 71.1 & 72.3 & 42.9 & 43.9 & 42.4 \\
        & Entropy & 46.4 & 46.9 & 46.8 & 72.6 & 72.6 & 73.2 & 70.9 & 71.4 & 72.4 & 42.4 & 41.8 & 42.9 \\
        & Margin & 46.3 & 46.5 & 49.2 & 72.5 & 73.5 & 74.1 & 71.0 & 71.5 & 71.8 & 43.4 & 43.4 & \underline{44.5} \\
        & Low-conf & 45.8 & 46.2 & 49.0 & 72.8 & 73.2 & 74.3 & 71.0 & 71.1 & 72.2 & 42.9 & 43.4 & \underline{44.5} \\
        \cmidrule(lr){2-14}
        & Ours (PSP) & \underline{47.6} & \underline{47.3} & \textbf{50.5} & \underline{74.3} & \underline{74.6} & \underline{75.0} & \underline{72.0} & \underline{72.5} & \underline{72.7} & \underline{44.5} & \underline{45.5} & \textbf{46.0} \\
        & Ours (PSP \& VRG) & \textbf{48.4} & \textbf{48.6} & \underline{50.3} & \textbf{74.9} & \textbf{75.2} & \textbf{75.3} & \textbf{72.8} & \textbf{72.7} & \textbf{73.4} & \textbf{45.5} & \textbf{46.6} & \textbf{46.0} \\
        \midrule
        \midrule

        \multirow{7}{*}{\textit{MMaDa}}
        & DDCoT & 34.1 & 34.0 & 34.1 & 55.7 & 55.8 & 55.5 & 56.2 & 56.4 & 55.8 & 35.0 & 34.5 & 36.1 \\
        & CCoT & 34.7 & 31.8 & 33.3 & 54.7 & 55.0 & 55.0 & 54.6 & 56.9 & 56.9 & 36.1 & 36.1 & 36.6 \\
        & Entropy & 34.1 & 33.6 & 34.3 & 56.2 & 55.7 & 55.5 & 56.0 & 56.7 & 57.3 & 36.1 & 35.6 & 35.0 \\
        & Margin & 34.5 & 33.8 & 33.8 & 55.6 & 55.5 & 55.8 & 56.0 & 56.8 & 57.1 & 34.0 & 35.6 & 34.5 \\
        & Low-conf & 33.7 & 33.8 & 34.6 & 56.1 & 56.0 & 55.7 & 56.4 & 56.7 & 57.3 & 35.6 & 35.0 & 34.5 \\
        \cmidrule(lr){2-14}
        & Ours (PSP) & \underline{35.6} & \underline{35.2} & \underline{35.8} & \underline{57.4} & \underline{57.5} & \underline{56.9} & \textbf{57.3} & \underline{57.5} & \underline{57.6} & \underline{37.7} & \underline{38.2} & \underline{37.1} \\
        & Ours (PSP \& VRG) & \textbf{36.3} & \textbf{36.6} & \textbf{36.4} & \textbf{59.9} & \textbf{59.1} & \textbf{58.1} & \underline{56.9} & \textbf{58.4} & \textbf{58.8} & \textbf{38.2} & \textbf{38.7} & \textbf{37.7} \\
        \bottomrule
    \end{tabular}
    \end{adjustbox}
\end{table*}

\subsection{Visual Reasoning Guidance}
\label{subsec:VRG}
In Section \ref{subsec:casestudy2}, we observe that dMLLMs exhibit weak dependency on visual evidence during reasoning. To address this issue, we extend the principle of Classifier-Free Guidance (CFG) \cite{ho2022classifier}, widely used in diffusion models, and propose Visual Reasoning Guidance (VRG). 

CFG amplifies the difference between the conditioned distribution $\epsilon_\theta(x_t \mid c)$ and the unconditioned distribution $\epsilon_\theta(x_t)$, encouraging the model to generate samples more faithful to the given condition $c$ (e.g., text prompt).  
Formally, it can be expressed as:
\begin{equation}
    \epsilon_{\text{cfg}} = \epsilon_{\text{uncond}} + s \, (\epsilon_{\text{cond}} - \epsilon_{\text{uncond}})
\end{equation}
where $s$ is the guidance scale controlling the influence strength of the condition.  

Similarly, in the context of dMLLM reasoning, we compute the conditional logits $\text{logits}_c$ (conditioned on the visual prompt $v$) and the unconditional logits $\text{logits}_u$ in parallel, and apply the following VRG formulation:

\begin{equation}
    \text{logits}_{\text{vrg}} = \text{logits}_u + (s_{\text{vrg}} + 1) \cdot (\text{logits}_c - \text{logits}_u)
\end{equation}
where $s_{\text{vrg}}$ denotes the visual guidance scale, which amplifies the influence of visual conditioning and strengthens the model’s reliance on visual information during reasoning process.

The guided logits are then used to sample tokens at each timestep for prediction.

Finally, by combining VRG and PSP, the confidence score $C_j^{i}$ used for the remasking strategy is defined as follows.

\begin{equation}
C_j^{i}
= \mathrm{softmax}\bigl(\text{logits}_{\mathrm{vrg}}\bigr)_j
= \frac{
    \exp\!\left(\text{logits}_{\mathrm{vrg}, j}\right)
}{
    \sum_{k \in \mathcal{C}}\exp\!\left(\text{logits}_{\mathrm{vrg}, k}\right)
}
\end{equation}

\begin{equation}
\tilde{C}_j^{i}
=
\frac{
    \exp\!\left(\text{logits}_{\mathrm{vrg}, j}\right)
}{
    \sum_{k \in \mathcal{C}}\exp\!\left(\text{logits}_{\mathrm{vrg}, k}\right)
}
\left[ 1 - \gamma (1 - \tau_i)\,\mathrm{rel}(j) \right]
\end{equation}
\section{Experiments}
\label{sec:experiments}

\subsection{Experimental Setup}
\label{subsec:experimental_setup}
\paragraph{Benchmarks.}
We evaluate our proposed method on four benchmarks: M3CoT \cite{m3cot}, ScienceQA \cite{scienceqa}, MMBench \cite{mmbench}, and V* Bench \cite{vstar}. M3CoT is designed to assess multimodal CoT reasoning across various domains, including science, mathematics, and commonsense. ScienceQA evaluates the model’s ability to integrate scientific knowledge with visual information for knowledge-based multimodal reasoning. MMBench includes diverse splits to evaluate a model’s general visual perception and visual reasoning capabilities. V* Bench is designed to focus on evaluating the model’s performance in high-resolution visual question answering tasks.

\paragraph{Models and Baselines.}
Our experiments are conducted based on two dMLLMs with strong reasoning capabilities: LaViDa-llada-reason \cite{lavida} and MMaDA-8B-MixCoT \cite{mmada}. Both models are additionally fine-tuned to generate rationales and are structured to generate progressive responses through an unmasking-based generation process. For a rigorous baseline setup, we include three remasking strategies: Low-confidence (Low-conf), Entropy, and Margin. We also incorporate two CoT-based methods, CCoT \cite{ccot} and DDCoT \cite{ddcot}, which have demonstrated strong performance in VLMs. Both CCoT and DDCoT employ the Low-conf strategy as the default remasking strategy of the dMLLMs.

\paragraph{Implementation Details.}
We evaluate our method by considering the speed-quality tradeoff, a key advantage of dMLLMs, across different generation lengths and diffusion steps ($L$ / $T$). The penalty coefficient $\gamma$ for PSP is fixed at 0.5, and the $s_{vrg}$ for VRG is also set to 0.5. To ensure strict experimental reproducibility, we do not apply temperature scaling and report results obtained using greedy decoding. The performance for each remasking strategy is presented in Table \ref{tab:remasking}, while Low-conf is used as the default remasking strategy in all other experiments. Details on prompt configurations and additional experimental setups are provided in the supplementary material.

\subsection{Main Results}
\label{subsec:main_results}

In Table \ref{tab:main}, existing methods originally designed for VLMs, such as DDCoT and CCoT, fail to achieve strong performance despite generating additional rationales. Notably, both DDCoT, which emphasizes a stepwise divide and conquer approach, and CCoT, which focuses on compositional visual reasoning, show limited improvement. This suggests that dMLLMs require a fundamentally different approach from AR-based VLMs. Neither DDCoT nor CCoT outperforms our methods or the representative unmasking strategies (Entropy, Margin, and Low-conf) under any generation length or diffusion steps.

In contrast, our method consistently outperforms all baselines, including both the remasking strategies (Entropy, Margin, and Low-conf) and VLM CoT methods (DDCoT / CCoT), across all experiments. First, our approach consistently delivers strong performance on both perception-focused benchmarks (MMBench and SQA-IMG) and reasoning-intensive benchmarks (M3CoT and V* Bench), demonstrating its robustness across different task types and difficulty levels. Second, despite the inherent difference in model reasoning capability between LaViDa and MMaDa, our method consistently improves performance across both models, showing that the effectiveness of our approach is not sensitive to model scale or base performance. Lastly, across all settings of generation length and diffusion steps, which are the key parameters of dMLLMs, our method achieves at least a 3\% improvement in accuracy. This demonstrates its general applicability and effectiveness across diverse dMLLM configurations.

From the perspective of efficiency, our method achieves superior performance under the $L$/$T$ = 64/32 compared to configurations using four times more diffusion steps. Given that dMLLMs inherently allow flexible control over the tradeoff between inference speed and response quality, our method effectively maximize the model’s capability. For example, LaViDa achieves 74.3\% accuracy using a generation length of 256 with the Low-conf strategy, whereas Low-conf + PSP \& VRG achieves 74.9\% accuracy with only a generation length of 64 in the MMBench benchmark. Similarly, on the same benchmark, MMaDa achieves 7.5\% higher accuracy with a generation length of 64 using our method compared to using a generation length of 256 with the Low-conf strategy. These results demonstrate both superior efficiency and reasoning quality.

\begin{table}[t]
  \centering
  \caption{LaViDa's ablation study results with generation length $L = 64$ / step $T = 32$. \textit{Low-conf} denotes the Low-confidence remasking strategy, and PSP and VRG are combined with Low-conf. \textbf{Bold} represents the best performance, and \underline{underline} indicates the second-best performance.}
  \label{tab:ablation}
  \resizebox{\columnwidth}{!}{
  \begin{tabular}{l|cccc}
    \toprule
    Method & \textbf{M$^3$CoT} & \textbf{MMBench} & \textbf{SQA-IMG} & \textbf{V* Bench} \\
    \midrule
    Low-conf & 45.8 & 72.8 & 71.0 & 42.9 \\
    Low-conf w/ PSP & 47.6 & 74.3 & 72.0 & 44.5 \\
    Low-conf w/ VRG & \underline{47.8} & \textbf{75.1} & \underline{72.1} & \underline{45.0} \\
    Low-conf w/ PSP \& VRG & \textbf{48.4} & \underline{74.9} & \textbf{72.8} & \textbf{45.5} \\
    \bottomrule
  \end{tabular}
  }
\end{table}

\begin{table}[t]
  \centering
  \caption{LaViDa's experimental results across different remasking strategies with generation length $L = 64$ / step $T = 32$. PSP and VRG are combined with each method.}
  \label{tab:remasking}
  \resizebox{\columnwidth}{!}{
  \begin{tabular}{l|cccc}
    \toprule
    Method & \textbf{M$^3$CoT} & \textbf{MMBench} & \textbf{SQA-IMG} & \textbf{V* Bench} \\
    \midrule
    Low-conf & 45.8 & 72.8 & 71.0 & 42.9 \\
    Low-conf w/ PSP \& VRG & 48.4 & 74.9 & 72.8 & 45.5 \\
    \midrule
    Entropy & 46.4 & 72.6 & 70.9 & 42.4 \\
    Entropy w/ PSP \& VRG & 48.0 & 74.9 & 72.6 & 46.0 \\
    \midrule
    Margin & 46.3 & 72.5 & 71.0 & 43.4 \\
    Margin w/ PSP \& VRG & 48.1 & 74.8 & 72.9 & 45.0 \\
    \bottomrule
  \end{tabular}
  }
\end{table}

\begin{figure}[t]
  \centering
  \includegraphics[width=1.0\linewidth]{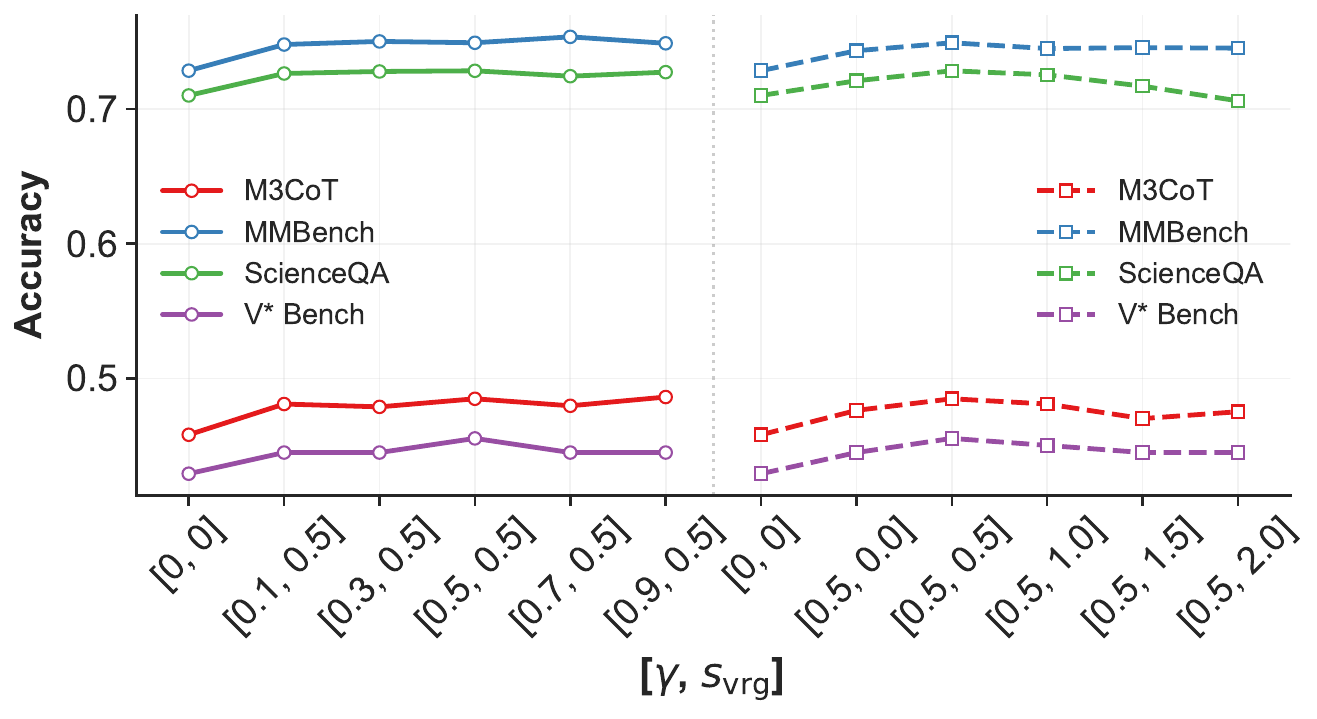}
  \caption{Performance comparison of LaViDa under different hyperparameter settings. On the horizontal axis, the notation \([x, y]\) represents the hyperparameter pair, where \(x\) denotes the penalty strength coefficient \(\gamma\) and \(y\) denotes the visual guidance scale $s_{vrg}$.}
  \label{fig:hyper}
\end{figure}

\subsection{Ablation \& Discussion}
\label{subsec:discussion}

\paragraph{Ablation Study.}
In Table \ref{tab:ablation}, we present the ablation study of PSP and VRG using LaViDa. Both PSP and VRG individually improve performance over the Low-conf (baseline), demonstrating that each method is effective on its own. Furthermore, applying them together yields the best performance across four benchmarks (M3CoT, MMBench, SQA-IMG, and V* Bench).


\paragraph{Generalization.}
Table \ref{tab:remasking} presents the results of applying PSP and VRG across three representative remasking strategies in dMLLMs. The remasking strategy is applied by replacing the confidence score in Equation \ref{eq:4} with entropy or margin. Experimental results show that our method consistently achieves significant performance improvements regardless of the remasking strategy. This suggests that PSP and VRG can operate in a plug-and-play manner with any remasking strategy, even as new strategies are introduced. In addition, Figure \ref{fig:hyper} shows the experimental results under different hyperparameter settings. Both PSP and VRG exhibit low sensitivity to hyperparameter variations while consistently outperforming the Low-conf (baseline) across all datasets. However, if $s_{\text{vrg}}$ is set excessively large, the model may ignore textual information and rely solely on visual information, so it is necessary to choose an appropriate value.

\paragraph{Efficiency.}
Table \ref{tab:time} presents the results of the time cost analysis. For a fair comparison, we use Low-conf as the default remasking strategy. The results show that DDCoT, which separates model inputs into reasoning and recognition for inference, incurs the highest time cost across all generation length and diffusion steps settings. In contrast, CCoT and PSP exhibit similar time costs to the Low-conf (baseline). Notably, PSP achieves better or comparable efficiency while consistently outperforming the Low-conf, DDCoT, and CCoT in all settings. VRG introduces a slightly higher time cost due to the computation of conditional and unconditional logits, but it remains highly competitive. It also shows no significant increase in time under the $L$/$T$ = 64/32 setting while still outperforming the $L$/$T$ = 256/128 configuration.

\begin{table}[t]
  \centering
  \caption{Average time consumption of LaViDa on the M3CoT validation set. The default remasking strategy used was Low-confidence, and DDCoT, CCoT, PSP, VRG, and PSP \& VRG were applied. The unit $s$ denotes seconds.}
  \label{tab:time}
  \resizebox{0.8\columnwidth}{!}{
  \begin{tabular}{l|ccc}
    \toprule
    Method & \textbf{64/32} & \textbf{128/64} & \textbf{256/128} \\
    \midrule
    Low-conf & 4.01s & 6.07s & 14.48s \\
    Low-conf w/ DDCoT & 6.03s & 9.99s & 25.95s \\
    Low-conf w/ CCoT & 4.05s & 6.09s & 14.59s \\
    Low-conf w/ PSP & 4.03s & 6.06s & 14.51s \\
    Low-conf w/ VRG & 4.73s & 8.46s & 22.05s \\
    Low-conf w/ PSP \& VRG & 4.65s & 8.43s & 22.29s \\
    \bottomrule
  \end{tabular}
  }
\end{table}

\begin{table}[t]
  \centering
  \caption{Comparison experimental results between PSP and L2R in LaViDa with generation length $L = 64$ / step $T = 32$. \textbf{Bold} indicates the better result between L2R \& VRG and PSP \& VRG, while \underline{underline} denotes the better result between L2R and PSP.}
  \label{tab:left2right}
  \resizebox{\columnwidth}{!}{
  \begin{tabular}{l|cccc}
    \toprule
    Method & \textbf{M$^3$CoT} & \textbf{MMBench} & \textbf{SQA-IMG} & \textbf{V* Bench} \\
    \midrule
    Low-conf & 45.8 & 72.8 & 71.0 & 42.9 \\
    Low-conf w/ L2R & 47.2 & 74.0 & 71.1 & 43.4 \\
    Low-conf w/ PSP & \underline{47.6} & \underline{74.3} & \underline{72.0} & \underline{44.5} \\
    Low-conf w/ L2R \& VRG & 47.6 & 74.6 & 71.3 & 44.5 \\
    Low-conf w/ PSP \& VRG & \textbf{48.4} & \textbf{74.9} & \textbf{72.8} & \textbf{45.5} \\
    \bottomrule
  \end{tabular}
  }
\end{table}

\begin{figure}[t]
  \centering
  \begin{subfigure}{\columnwidth}
    \includegraphics[width=\textwidth]{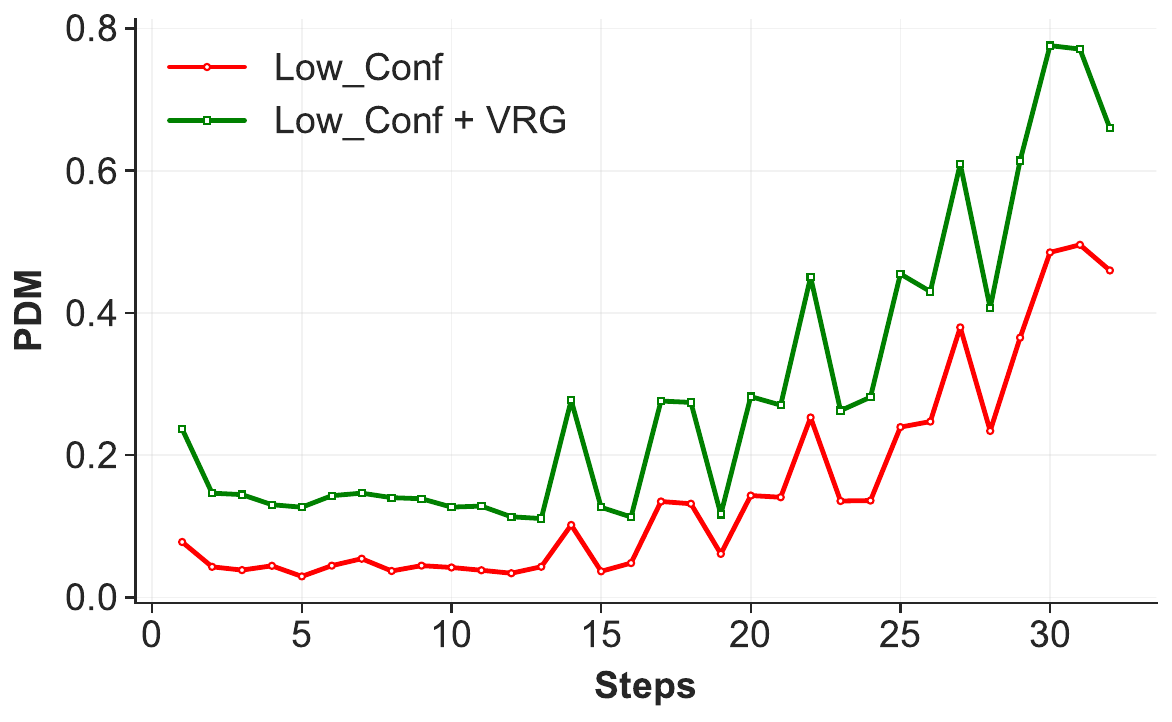}
    \caption{LaVida with generation length $L$ = 64 / step $T$ = 32}
    \label{fig:pdm_64}
  \end{subfigure}
  \vspace{3mm} 
  \begin{subfigure}{\columnwidth}
    \includegraphics[width=\textwidth]{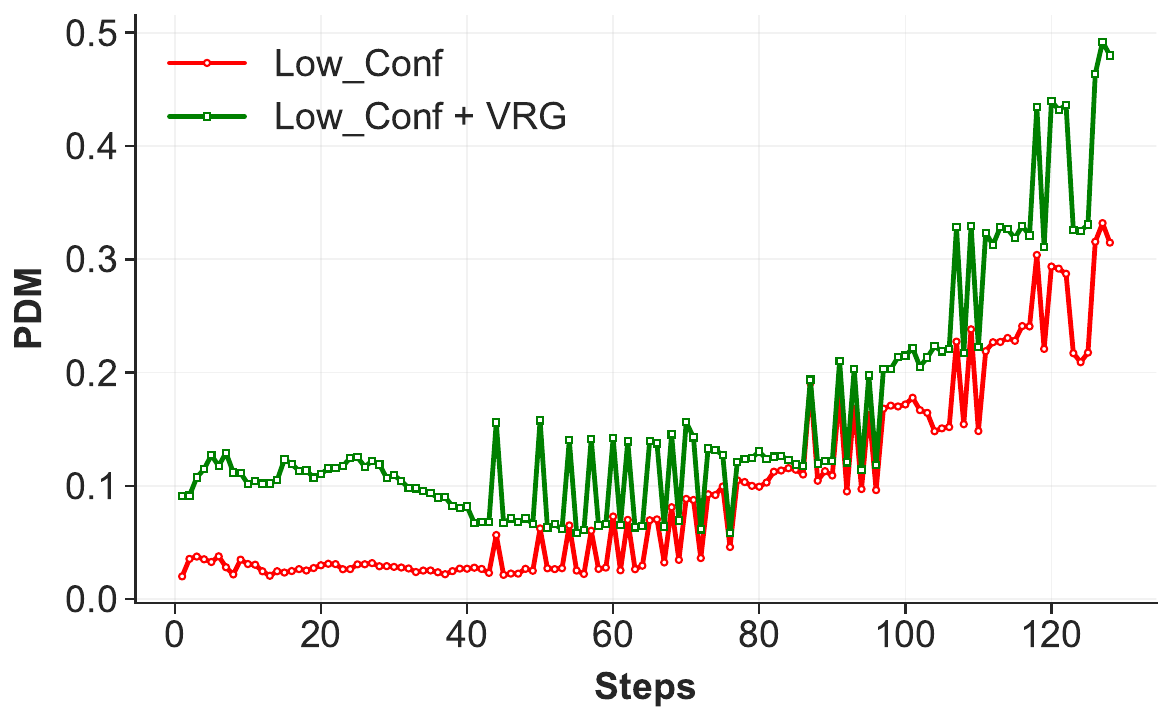}
    \caption{LaVida with generation length $L$ = 256 / step $T$ = 128}
    \label{fig:pdm_256}
  \end{subfigure}
  \caption{Comparison of PDM measurements on the M3CoT validation set between Low-conf and Low-conf + VRG using LaViDa. the visual guidance scale $s_{vrg}$ = 0.5.}
  \label{fig:pdm_comparison}
\end{figure}

\paragraph{Discussion.}
To compare PSP with AR generation, we conduct an experiment in which dLLMs generate tokens in a Left-to-Right (L2R) manner by unmasking the leftmost position at each timesteps. For example, when $L = 64$ and step $T = 32$, two leftmost remaining masked tokens are unmasked at each timesteps. As shown in Table \ref{tab:left2right}, PSP consistently outperforms the L2R method across all datasets. Moreover, when combined with VRG, PSP \& VRG achieve higher performance than L2R \& VRG in every dataset.

Figure \ref{fig:pdm_comparison} shows how the average PDM vary over timesteps on the M3CoT validation set when using LaViDa with Low-conf (base) and VRG. In the Low-conf, PDM remains relatively low during the early timesteps, indicating that the model initially relies on limited visual information. When VRG is applied, PDM is consistently higher across timesteps, demonstrating that visual grounding is effectively strengthened throughout the generation process.

\section{Conclusion}
\label{sec:conclusion}

In this paper, we analyze the reasoning process of dMLLMs and propose a novel method to address their limitations. Our analysis shows that dMLLMs exhibit two major issues: (1) Early Answer Generation, where the model tends to generate the final answer prematurely before undergoing sufficient reasoning steps, and (2) low dependency on visual prompts during the early timesteps. These findings suggest that, due to the parallel token restoration mechanism inherent to dMLLMs, it is challenging to directly apply reasoning enhancement methods originally designed for AR models.
To mitigate these issues, we propose Position \& Step Penalty (PSP), which encourages dMLLMs to perform progressive reasoning throughout the diffusion process rather than generating final answers too early. In addition, we introduce Visual Reasoning Guidance (VRG) to strengthen the model’s reliance on visual evidence. As a result, we achieve improvements in both efficiency and reasoning quality without additional training.
Our study provides a promising direction for enabling dMLLMs to approach the reasoning capabilities of AR VLMs, and we expect it will further promote the development and utilization of dMLLMs in future research.

\section*{Acknowledgements}
This work was supported by the Institute of Information and communications Technology Planning and evaluation (IITP) grant (No.RS-2025-25422680, No. RS-2020-II201373), and the National Research Foundation of Korea (NRF) grant (No. RS-2025-00520618) funded by the Korean Government (MSIT).

{
    \small
    \bibliographystyle{ieeenat_fullname}
    \bibliography{main}
}

\clearpage
\setcounter{page}{1}
\maketitlesupplementary

\begin{figure}[t]
  \centering
  \begin{subfigure}{\columnwidth}
    \includegraphics[width=\textwidth]{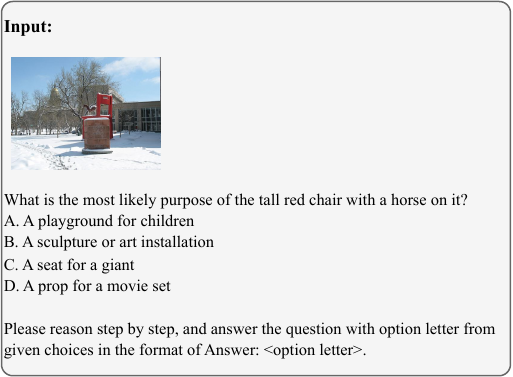}
    \caption{LaViDa}
    \label{fig:lavida_prompt}
  \end{subfigure}
  \vspace{3mm} 
  \begin{subfigure}{\columnwidth}
    \includegraphics[width=\textwidth]{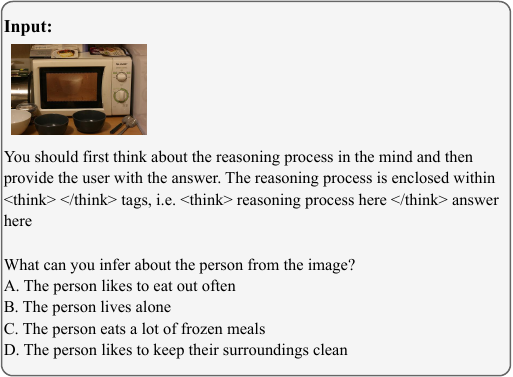}
    \caption{MMaDa}
    \label{fig:mmada_prompt}
  \end{subfigure}
  \caption{Example of a reasoning prompt based on each model’s reference implementation.}
  \label{fig:prompt}
\end{figure}

\section{Prompting Details}
\label{sec:prompting}
This section provides additional details about our prompting used for diffusion-based multimodal reasoning. Following LaViDa and MMaDa, we adopt the think prompt to encourage structured, step-by-step reasoning during generation. The think prompt guides the model to first produce intermediate reasoning before generating the final answer, thereby improving interpretability and mitigating early answer generation. Figure \ref{fig:prompt} shows the complete think prompt templates used in all our experiments.

\section{Additional Results}
\label{sec:add_results}

\subsection{Additional Analysis}
\label{subsec:add_anal}

\begin{figure}[t]
  \centering
  \begin{subfigure}{\columnwidth}
    \includegraphics[width=\textwidth]{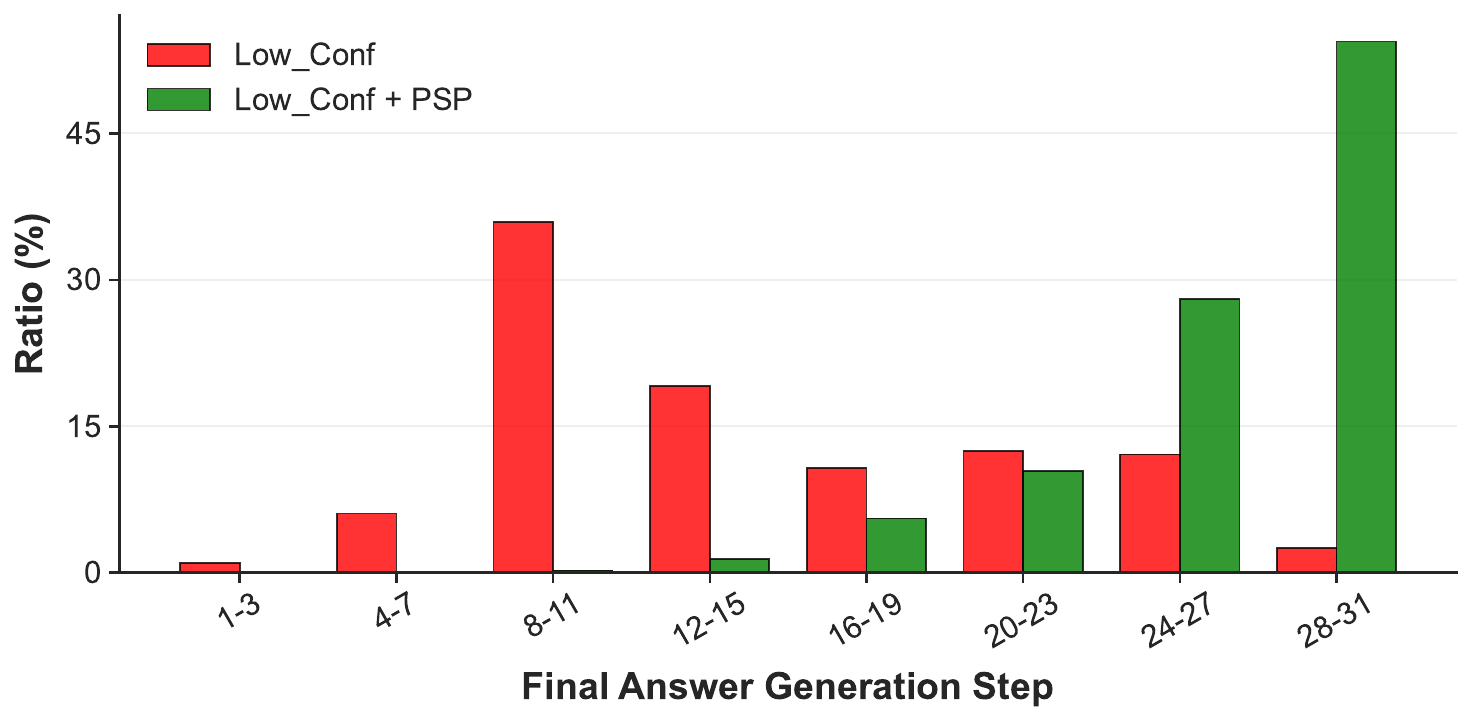}
    \caption{Generation length $L$ = 64 / diffusion step $T$ =32}
    \label{fig:answer_steps_64_mmada}
  \end{subfigure}
  \vspace{0mm} 
  \begin{subfigure}{\columnwidth}
    \includegraphics[width=\textwidth]{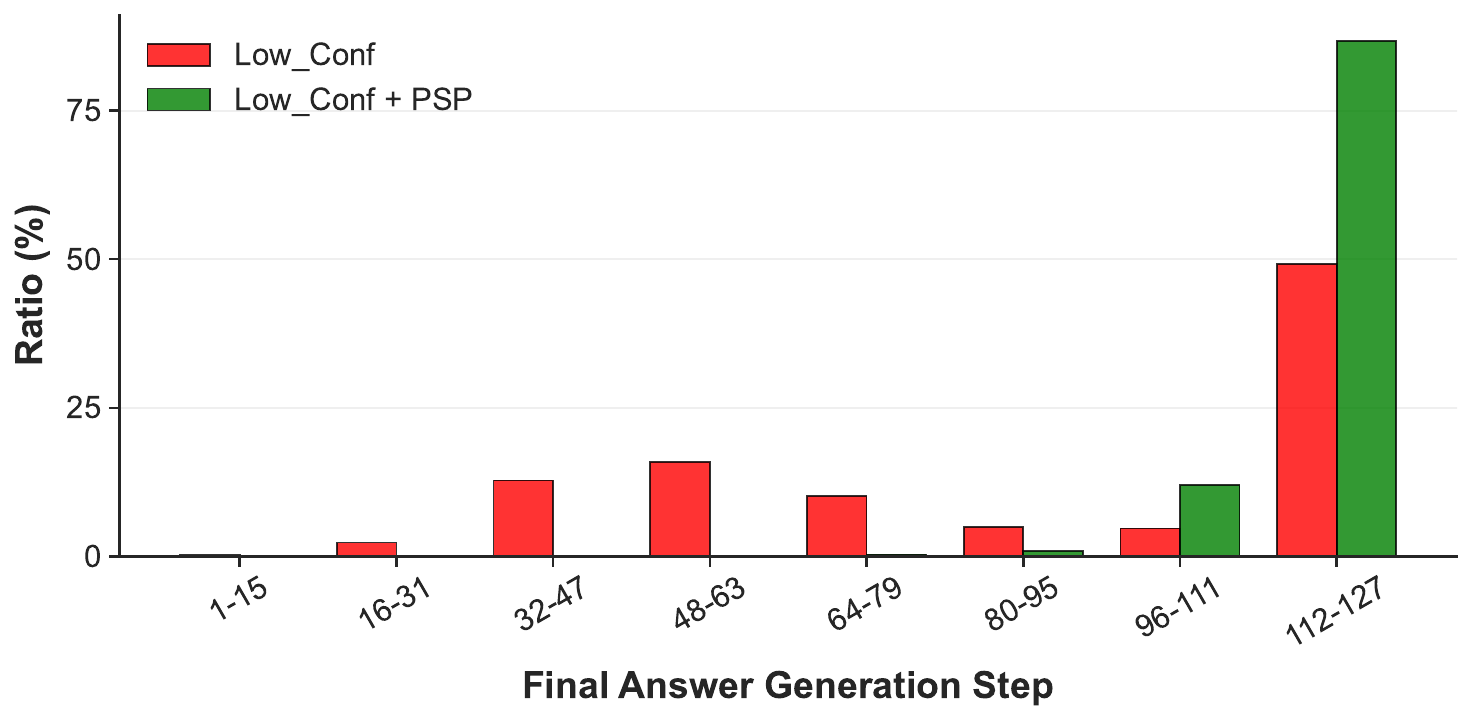}
    \caption{Generation length $L$ = 256 / diffusion step $T$ = 128}
    \label{fig:answer_steps_256_mmada}
  \end{subfigure}
  \caption{Results of the final answer generation step on the M3CoT validation set using MMaDa. The default remasking strategy is Low-confidence.}
  \label{fig:answer_step_mmada}
\end{figure}

\begin{figure}[t]
  \centering
  \begin{subfigure}{\columnwidth}
    \includegraphics[width=\textwidth]{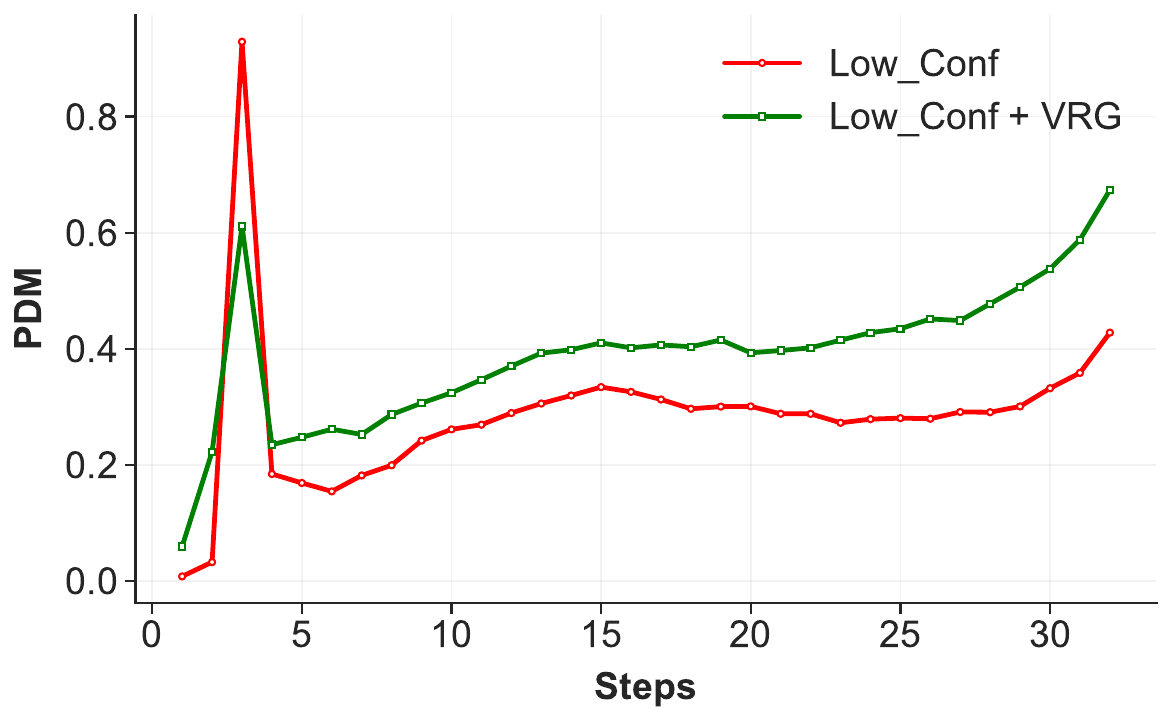}
    \caption{MMaDa with generation length $L$ = 64 / step $T$ = 32}
    \label{fig:pdm_64_mmada}
  \end{subfigure}
  \vspace{3mm} 
  \begin{subfigure}{\columnwidth}
    \includegraphics[width=\textwidth]{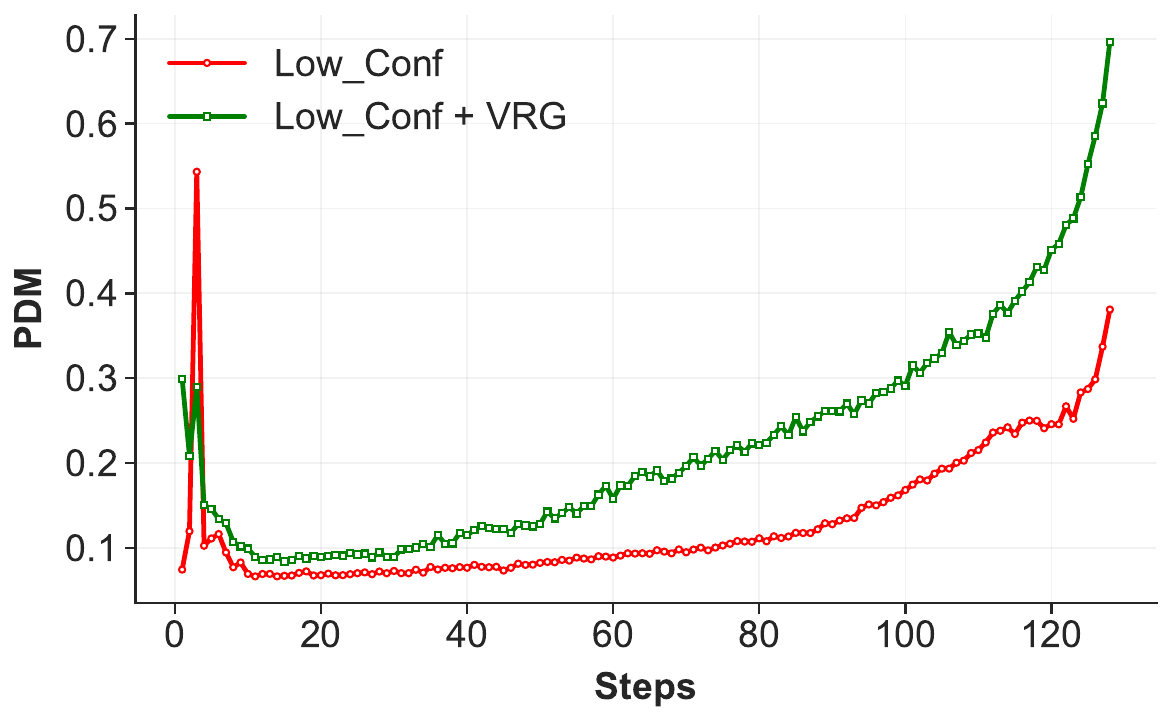}
    \caption{MMaDa with generation length $L$ = 256 / step $T$ = 128}
    \label{fig:pdm_256_mmada}
  \end{subfigure}
  \caption{Comparison of PDM measurements on the M3CoT validation set between Low-conf and Low-conf + VRG using MMaDa.}
  \label{fig:pdm_step_comparison_mmada}
\end{figure}

In this subsection, we conduct additional experiments on MMaDa. The results corresponding to Observation 1 and Observation 2 are presented in Figure \ref{fig:answer_step_mmada} and Figure \ref{fig:pdm_step_comparison_mmada}, respectively. As shown in Figure \ref{fig:answer_step_mmada}, MMaDa exhibits a clear Early Answer Generation, similar to what we observe in LaViDa. The model frequently generates the final answer at very early timesteps, indicating premature answer determination before sufficient reasoning. However, when PSP is applied, the distribution shifts toward later timesteps, encouraging more gradual reasoning and effectively mitigating early answer generation.

Likewise, Figure \ref{fig:pdm_step_comparison_mmada} shows that MMaDa demonstrates low visual dependence during early timesteps, again consistent with the property seen in LaViDa. The model begins to meaningfully incorporate visual information only in later steps. Applying VRG significantly increases visual dependency across the diffusion process, reinforcing visual grounding and enabling earlier and more consistent use of visual evidence. These results indicate that PSP and VRG improve reasoning progression and visual grounding not only in LaViDa but also in MMaDa, demonstrating their general applicability across different dMLLMs.

\begin{figure}[t]
  \centering
  \begin{subfigure}{\columnwidth}
    \includegraphics[width=\textwidth]{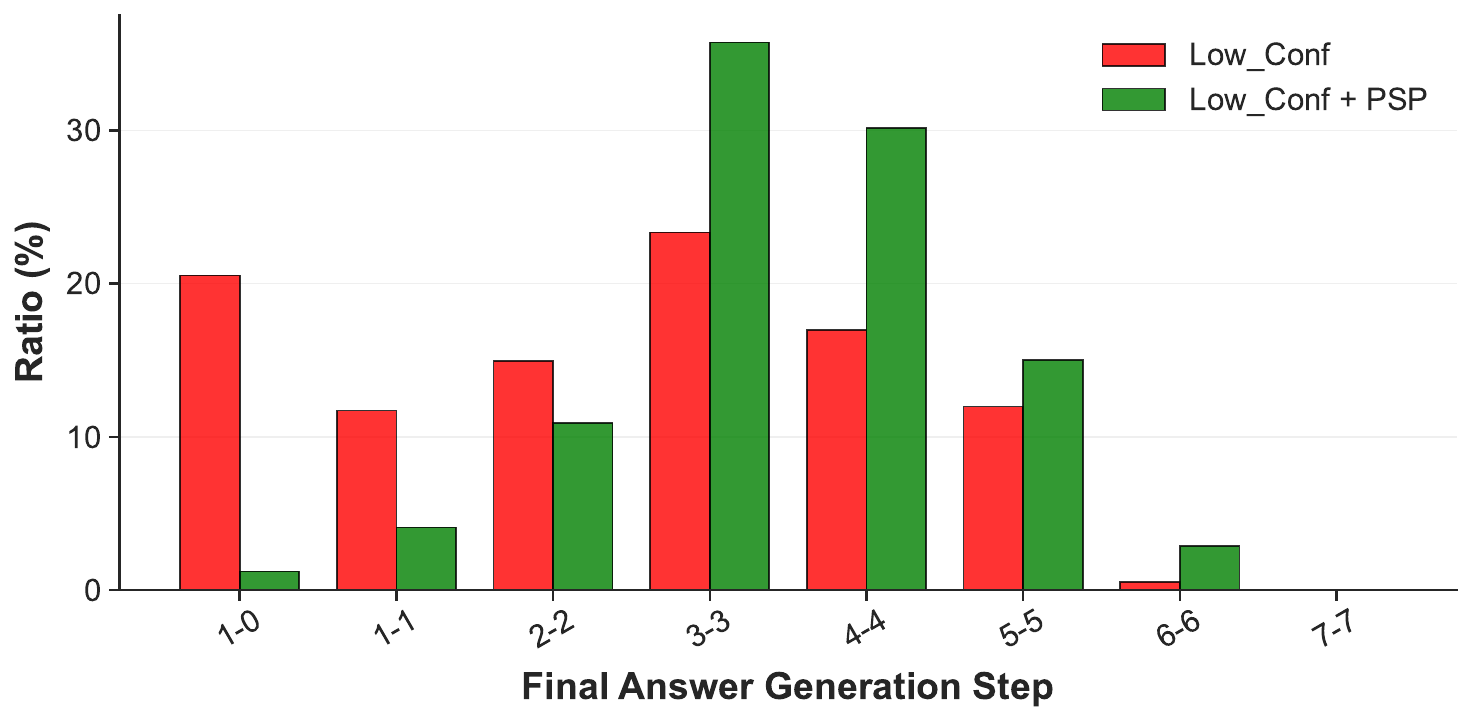}
    \caption{Generation length $L$ = 64 / diffusion step $T$ =8}
    \label{fig:answer_steps_64_8}
  \end{subfigure}
  \vspace{0mm} 
  \begin{subfigure}{\columnwidth}
    \includegraphics[width=\textwidth]{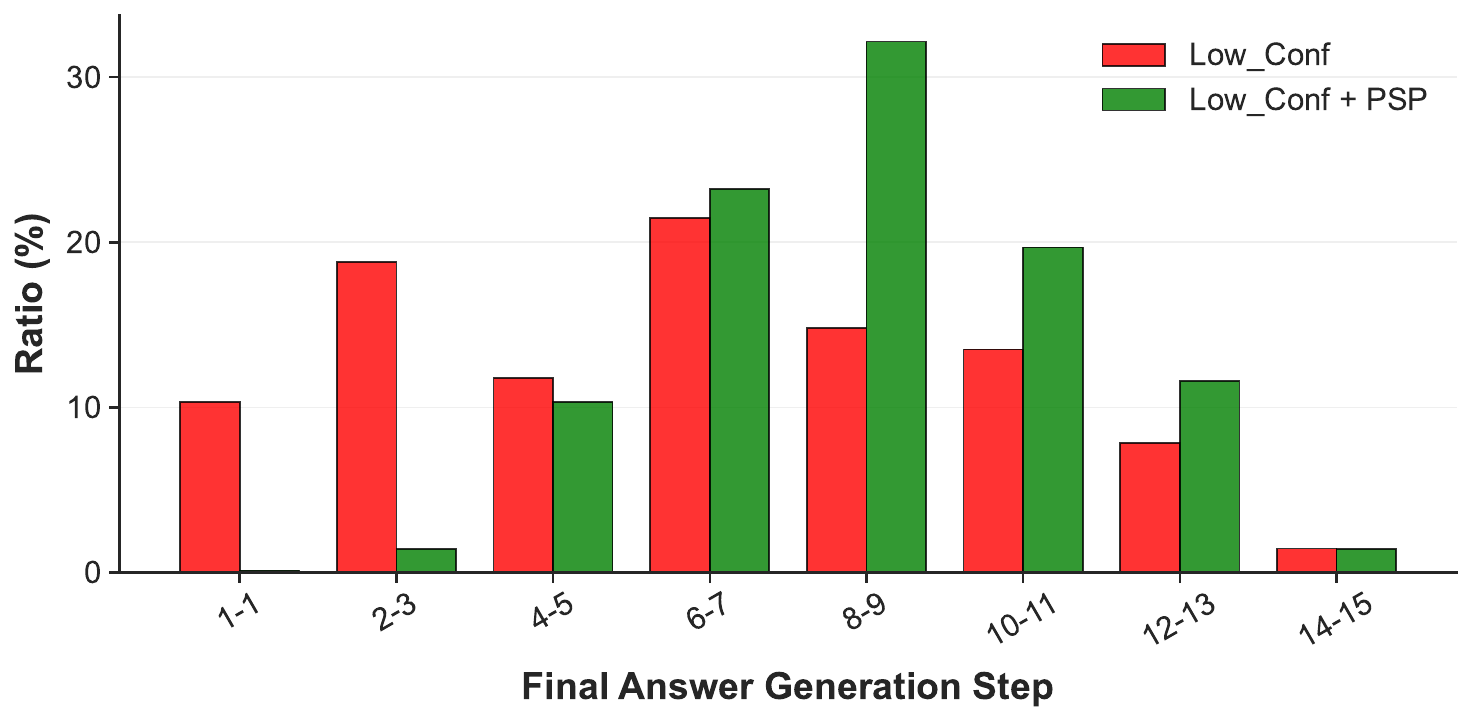}
    \caption{Generation length $L$ = 64 / diffusion step $T$ =16}
    \label{fig:answer_steps_64_16}
  \end{subfigure}
  \vspace{0mm} 
  \begin{subfigure}{\columnwidth}
    \includegraphics[width=\textwidth]{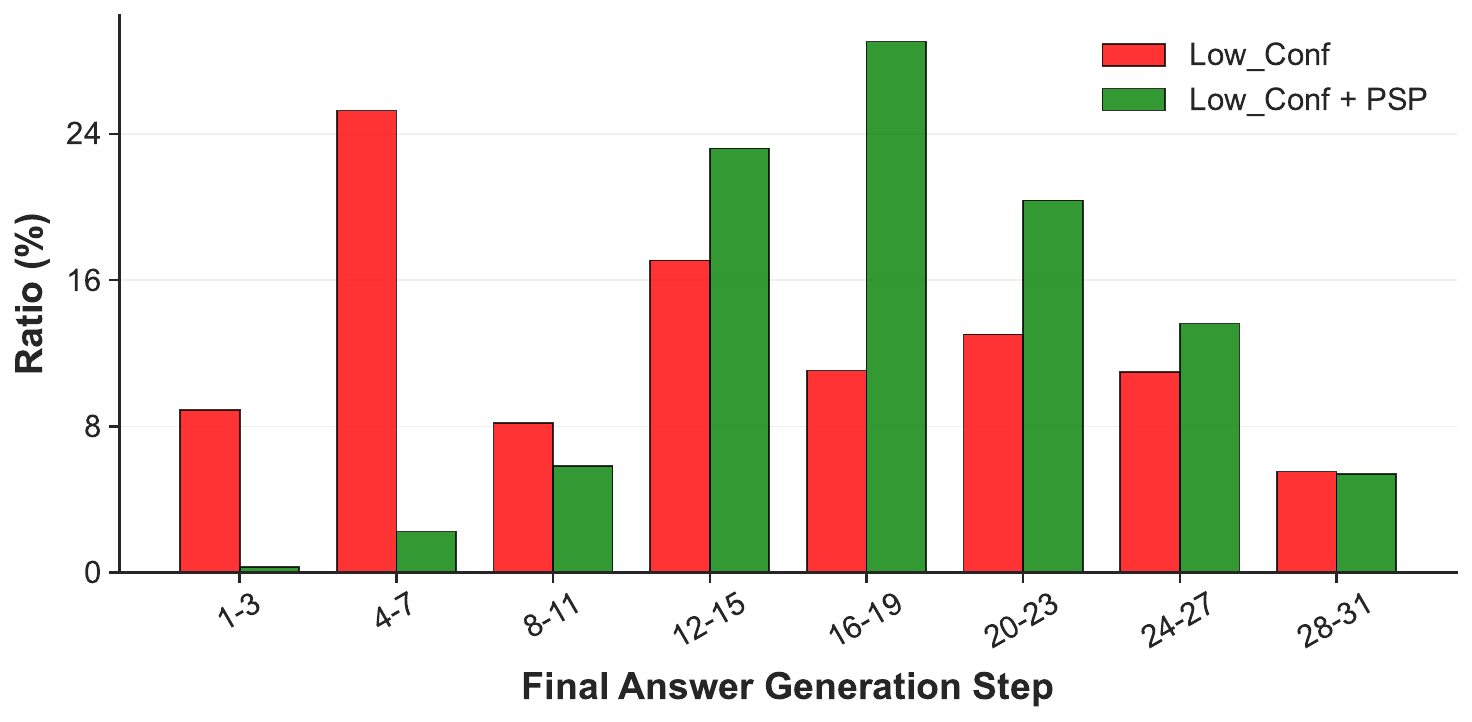}
    \caption{Generation length $L$ = 64 / diffusion step $T$ = 32}
    \label{fig:answer_steps_64_32}
  \end{subfigure}
  \vspace{0mm} 
  \caption{Results of the final answer generation step on the M3CoT validation set using LaViDa, evaluated across different diffusion steps $T$.}
  \label{fig:answer_step_lavida}
\end{figure}

\subsection{Analysis on Varying Diffusion Steps}
\label{subsec:anal_steps}
\begin{figure}[t]
  \centering
\begin{subfigure}{\columnwidth}
    \includegraphics[width=\textwidth]{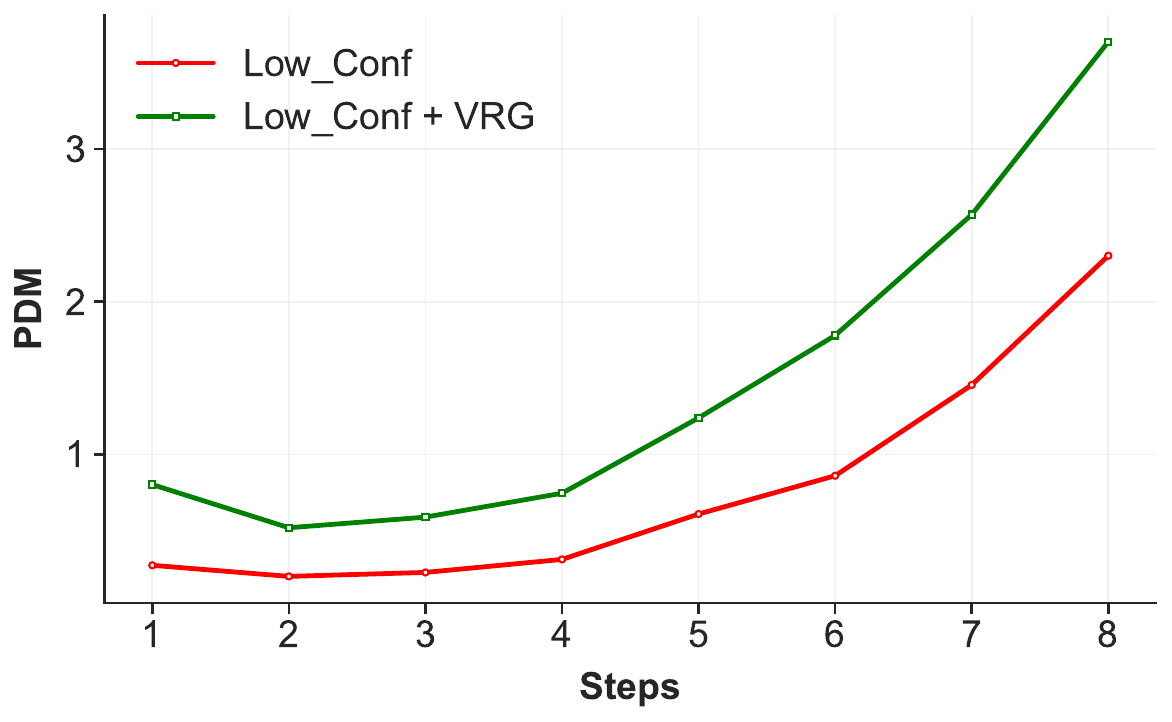}
    \caption{LaVida with generation length $L$ = 64 / step $T$ = 8}
    \label{fig:pdm_64_8}
  \end{subfigure}
  \vspace{3mm} 
  \begin{subfigure}{\columnwidth}
    \includegraphics[width=\textwidth]{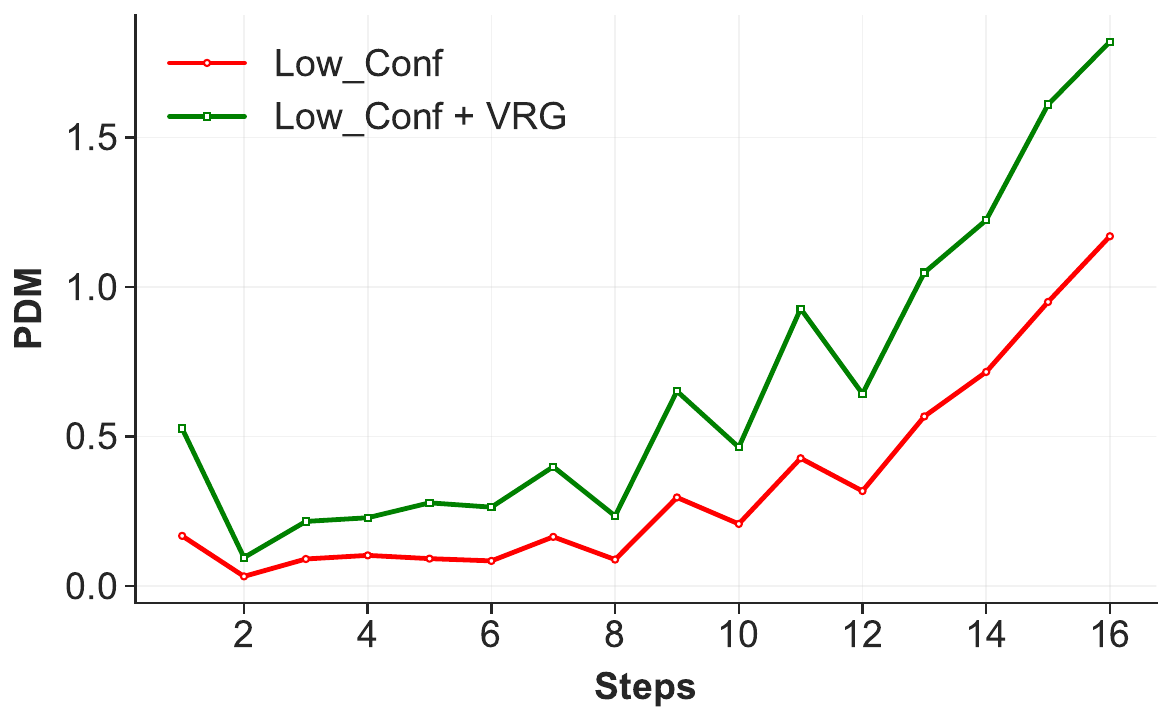}
    \caption{LaVida with generation length $L$ = 64 / step $T$ = 16}
    \label{fig:pdm_64_16}
  \end{subfigure}
  \vspace{3mm} 
  \begin{subfigure}{\columnwidth}
    \includegraphics[width=\textwidth]{figure/pdm_64.pdf}
    \caption{LaVida with generation length $L$ = 64 / step $T$ = 32}
    \label{fig:pdm_64_32}
  \end{subfigure}
  \caption{Comparison of PDM measurements on the M3CoT validation set between Low-conf and Low-conf + VRG using LaViDa, evaluated across different diffusion steps $T$.}
  \label{fig:pdm_step_comparison_lavida}
\end{figure}

We further analyze LaViDa by fixing the generation length to $L=64$ and varying the number of diffusion steps across three settings: $T=8$, $T=16$, and $T=32$. The corresponding results for Early Answer Generation and visual prompt dependency are presented in Figure \ref{fig:answer_step_lavida} and Figure \ref{fig:pdm_step_comparison_lavida}, respectively.

As shown in Figure \ref{fig:answer_step_lavida}, LaViDa consistently exhibits strong Early Answer Generation when operating under low diffusion steps. With $T=8$, the model frequently generates the final answer within the first few steps, indicating premature answer formation. Increasing the number of diffusion steps to $T=16$ and $T=32$ shifts the answer-generation distribution toward later timesteps, but the early-answer tendency remains visible. Applying PSP effectively suppresses this behavior across all three settings, pushing the answer generation toward later stages and encouraging more gradual reasoning even when the diffusion schedule is highly constrained.

\begin{table*}[htbp]
    \centering
    \caption{Comparison of dMLLM under varying numbers of diffusion steps $T$ with a fixed generation length $L=64$. X/Y denotes the generation length $L$ and step size $T$, respectively.}
    \label{tab:additional_steps}
    \begin{adjustbox}{max width=\textwidth}
    \begin{tabular}{l|l|cccc|cccc|cccc}
        \toprule
        \multirow{2}{*}{\textbf{Model}} & \multirow{2}{*}{\textbf{Method}} &
        \multicolumn{4}{c}{\textbf{M$^3$CoT}} & 
        \multicolumn{4}{c}{\textbf{MMBench}} & 
        \multicolumn{4}{c}{\textbf{SQA-IMG}} \\
        \cmidrule(lr){3-6} \cmidrule(lr){7-10} \cmidrule(lr){11-14}
        & & 64/8 & 64/16 & 64/32 & 64/64 & 64/8 & 64/16 & 64/32 & 64/64 & 64/8 & 64/16 & 64/32 & 64/64  \\
        \midrule

        \multirow{4}{*}{\textit{LaViDa}} 
        & Entropy & 46.2 & 46.3 & 46.4 & 47.0 & 72.7 & 72.8 & 72.6 & 72.7 & 70.8 & 70.9 & 70.9 & 71.2 \\
        & Margin & 46.4 & 46.5 & 46.3 & 46.9 & 72.3 & 72.9 & 72.5 & 73.0 & 71.3 & 71.0 & 71.0 & 71.5 \\
        & Low-conf & 45.3 & 45.7 & 45.8 & 46.4 & 72.6 & 72.9 & 72.8 & 72.8 & 71.1 & 70.6 & 71.0 & 71.3 \\
        \cmidrule(lr){2-14}
        & Ours & \textbf{47.9} & \textbf{47.7} & \textbf{48.4} & \textbf{48.5} & \textbf{74.4} & \textbf{74.9} & \textbf{74.9} & \textbf{74.8} & \textbf{72.1} & \textbf{72.3} & \textbf{72.8} & \textbf{72.7} \\
        
        \bottomrule
    \end{tabular}
    \end{adjustbox}
\end{table*}

Similarly, Figure \ref{fig:pdm_step_comparison_lavida} shows the visual prompt dependency (PDM) under the same settings. When using the default remasking strategy, LaViDa shows weak visual dependence at early steps across all values of $T$. The PDM gradually increases as the diffusion progresses, but the early-stage visual grounding remains minimal. With the application of VRG, the PDM curves consistently rise across all diffusion steps, demonstrating substantially stronger and earlier integration of visual information. Notably, the improvement becomes more pronounced as $T$ increases, showing that VRG enhances visual grounding regardless of the diffusion schedule.

These findings indicate that the reasoning characteristics observed in the main paper, namely early answer generation and weak early visual grounding, persist across different diffusion step configurations. Furthermore, PSP and VRG reliably address these issues under all tested settings, confirming their robustness even when the diffusion budget is significantly limited.

\subsection{Performance on Diffusion Steps}
\label{subsec:steps_vary}

We further investigate the influence of the number of diffusion steps $T$ on the reasoning performance. To isolate the effect of $T$, we fix the generation length to $L=64$ for all experiments and vary only the number of diffusion steps. Across all evaluated remasking strategies, we observe a trend that a reduction in the number of diffusion steps results in a consistent decrease in overall accuracy. Despite the reduction in performance at small $T$, our method remains highly effective. Across all tested values of $T$, our approach achieves state-of-the-art performance under very constrained diffusion schedules, as shown in Table \ref{tab:additional_steps}. These results indicate that our method not only improves reasoning quality but also provides robustness to the choice of diffusion steps, enabling stable performance across a wide range of inference-time configurations.

\subsection{Experiments on Long-form Answers}
\label{subsec:results_longform}

We further investigate whether the issue described in Observation 1 also arises in long-form answer settings. To this end, we conduct the same analysis on LLaVA-Bench COCO with long-form answers. By enforcing the output format \texttt{<Mask><Mask> Answer: <Mask><Mask>}, we define the first timestep at which 75\% of the tokens following “Answer:” are filled as the answer generation point. As shown in Figure \ref{fig:onecol}, Early Answer Generation is consistently observed.

\setcounter{figure}{11}
\begin{figure}[t]
    \setlength{\abovecaptionskip}{2pt}
    \setlength{\belowcaptionskip}{0pt}
    \centering
    \includegraphics[width=\linewidth]{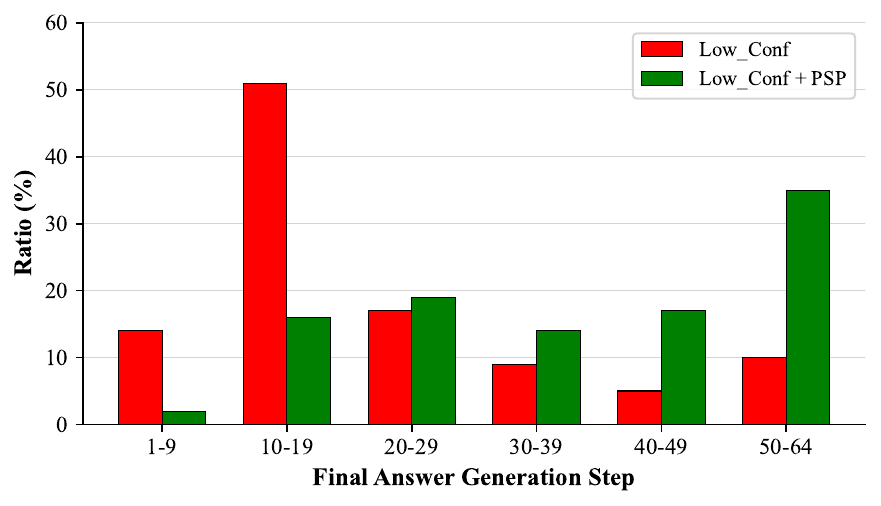}
    \caption{Observation 1 for LLaVA-Bench COCO.}
    \label{fig:onecol}
\end{figure}

\setcounter{table}{6}
\begin{table*}[t]
\setlength{\abovecaptionskip}{2pt}
\setlength{\belowcaptionskip}{0pt}
\centering
\small
\begin{tabular}{lcccccc}
\toprule
 & MME Exist. $\uparrow$ & MME Count $\uparrow$ & MME Pos. $\uparrow$ & MME Color $\uparrow$ & MME Total $\uparrow$ & LLaVA-Bench $\uparrow$\\
\midrule
Low Conf. & 183.33 & 133.33 & 86.67 & 141.67 & 545.00 & 16.5 \\
CCoT      & 185.00 & 126.67 & 90.55 & 148.33 & 550.55 & 17.2 \\
DDCoT     & 176.67 & 137.22 & 81.67 & 149.17 & 544.73 & 15.4 \\
PSP \& VRG      & \textbf{187.00} & \textbf{143.33} & \textbf{91.67} & \textbf{150.00} & \textbf{570.00} & \textbf{20.0} \\
\bottomrule
\end{tabular}
\caption{Experimental results on the MME benchmark for LaViDa with the 64/128 setting.}\label{tab:rebuttal1}
\end{table*}

\subsection{Results on Complex Metrics / Datasets}
\label{subsec:results_complex}

To evaluate a broader range of perception and cognition abilities beyond simple accuracy-based measurements, we conduct experiments on the MME benchmark and LLaVA-Bench. We additionally evaluate image perception and cognition using MME and conduct experiments on LLaVA-Bench with descriptive responses. As shown in Table \ref{tab:rebuttal1}, PSP \& VRG consistently achieves the best performance across all metrics.

\end{document}